\documentclass[runningheads]{llncs}
\usepackage{graphicx}
\usepackage[export]{adjustbox}
\usepackage{comment}
\usepackage{amsmath,amssymb} % define this before the line numbering.
\usepackage{booktabs}
\usepackage{multirow}
\usepackage{siunitx}
\usepackage{multicol}
\usepackage{leftidx}
\usepackage{graphics}
\usepackage{algorithm} 
\usepackage{algpseudocode}
\usepackage{authblk}
\usepackage{microtype}
\usepackage{amssymb}% http://ctan.org/pkg/amssymb
\usepackage{pifont}% http://ctan.org/pkg/pifont
\usepackage{subcaption}
\newcommand{\xmark}{\ding{55}}%

\begin{document}
% \linenumbers
\pagestyle{headings}
\mainmatter
\def\ECCVSubNumber{5809}  % Insert your submission number here

\title{SMART: Simultaneous Multi-Agent Recurrent Trajectory Prediction} % Replace with your title

\begin{comment}
\titlerunning{ECCV-20 submission ID \ECCVSubNumber} 
\authorrunning{ECCV-20 submission ID \ECCVSubNumber} 
\author{Anonymous ECCV submission}
\institute{Paper ID \ECCVSubNumber}
\end{comment}
%******************

% \begin{comment}
\titlerunning{SMART: Simultaneous Multi-Agent Recurrent Trajectory Prediction}
\author{Sriram N N\inst{1} \and
Buyu Liu\inst{1} \and
Francesco Pittaluga\inst{1} \and
Manmohan Chandraker \inst{1,2}}
\authorrunning{S. N N, et al.}
\institute{NEC Laboratories America \and
UC San Diego
}
% \end{comment}
%******************
\maketitle

\begin{abstract}

We propose advances that address two key challenges in future trajectory prediction: (i) multimodality in both training data and predictions and (ii) constant time inference regardless of number of agents. Existing trajectory predictions are fundamentally limited by lack of diversity in training data, which is difficult to acquire with sufficient coverage of possible modes. Our first contribution is an automatic method to simulate diverse trajectories in the top-view. It uses pre-existing datasets and maps as initialization, mines existing trajectories to represent realistic driving behaviors and uses a multi-agent vehicle dynamics simulator to generate diverse new trajectories that cover various modes and are consistent with scene layout constraints. Our second contribution is a novel method that generates diverse predictions while accounting for scene semantics and multi-agent interactions, with constant-time inference independent of the number of agents. We propose a convLSTM with novel state pooling operations and losses to predict scene-consistent states of multiple agents in a single forward pass, along with a CVAE for diversity. We validate our proposed multi-agent trajectory prediction approach by training and testing on the proposed simulated dataset and existing real datasets of traffic scenes. In both cases, our approach outperforms SOTA methods by a large margin, highlighting the benefits of both our diverse dataset simulation and constant-time diverse trajectory prediction methods.

\end{abstract}

\section{Introduction}
\begin{figure}[t]
    \centering
    \includegraphics[width=0.95\linewidth]{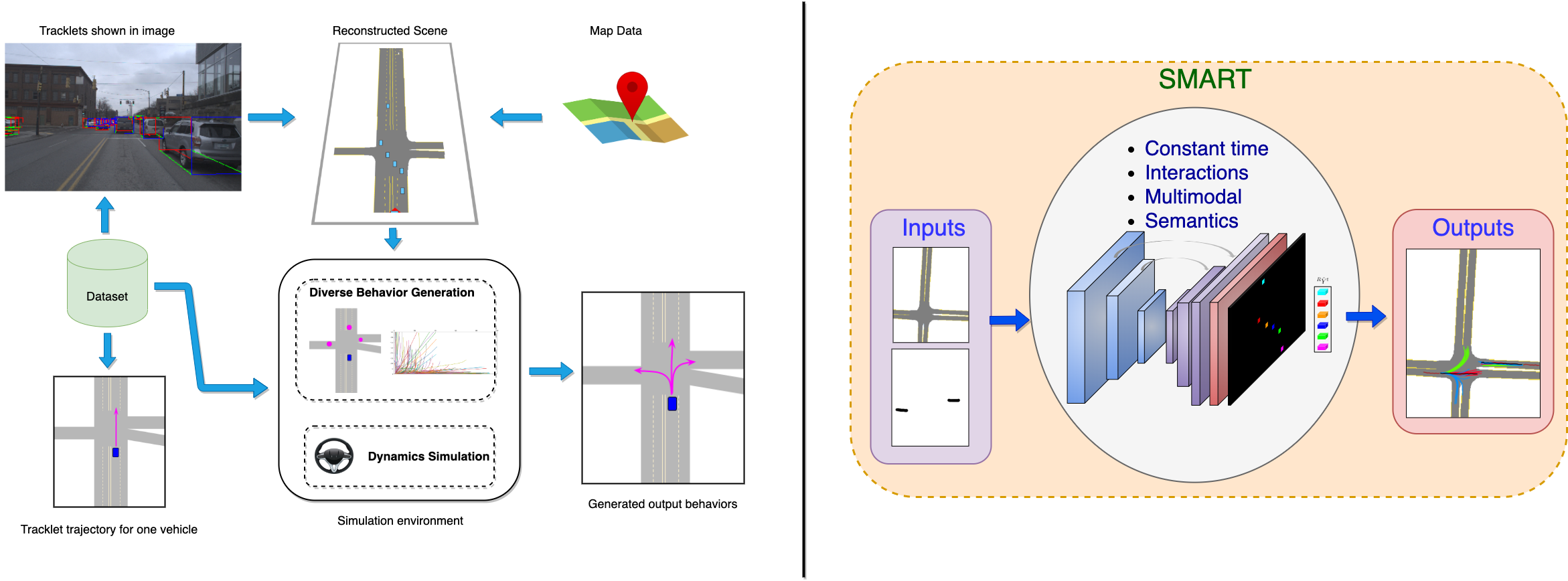}
    % \begin{tabular}{cc}
    %     \includegraphics[width=0.5\linewidth, height=0.3\linewidth]{./images/teaser_nec_pp.png} &  
    %     % \includegraphics[width=0.5\linewidth, height=0.3\linewidth]{./images/teaser_smart_2.png}\\
    % \end{tabular}
    \caption{Left: Given the map and tracklets, we propose to reconstruct the real world scene in top-view and simulate diverse behaviors for multiple agents w.r.t. scene context. Right: The proposed SMART algorithm that is able to generate context aware and multimodal trajectories for multiple agents. }
    \label{fig:teaser}
\end{figure}

The ability to reason about the future states of multiple agents in a scene is an important task for applications that seek vehicle autonomy. Ideally, a prediction framework should have three properties. First, it must be able to predict multiple plausible trajectories in the dominant modes of motion. Second, these trajectories should be consistent with the scene semantics. Third, it is attractive for several applications if constant-time prediction can be achieved regardless of the number of agents in the scene. In this paper, we propose dataset creation and future prediction methods that help achieve the above three properties (Figure \ref{fig:teaser}).

A fundamental limitation for multimodal trajectory prediction is the lack of training data with a diverse enough coverage of the possible motion modes in a scene. Our first main contribution is a simulation strategy to recreate driving scenarios from real world data, which generates multiple driving behaviors to obtain diverse trajectories for a given scene. We construct a graph-based simulation environment that leverages scene semantics and maps to execute realistic vehicle behaviors in the top-view. We sample reference velocity profiles from trajectories executing similar maneuvers in the real world data. Then we use a variant of the Intelligent Driver Model \cite{Treiber2000CongestedTS,highway-env} to model the dynamics of vehicle driving patterns and introduce lane-change decisions for simulated vehicles based on MOBIL \cite{mobil}. We show that training with our simulated datasets leads to large improvements in prediction outcomes compared to the real data counterparts that are comparatively limited in both scale and diversity reflected by a Wasserstein metric.

Several recent works consider deep networks for trajectory prediction for humans \cite{social_gan,social_ways_cvprW19,kosaraju2019socialbigat,social_lstm,sophiegan} and vehicles \cite{DEISRE,INFER,rulesofroad_cvpr19,traphic_cvpr19,MATF,Rhinehart_2018_ECCV}. Usually, they consider interactions among multiple agents, but still operate on single agent basis at inference time, requiring one forward pass for each agent in the scene. Vehicle motions are stochastic and depending on their goals, obtaining multimodal predictions for individual vehicles that are consistent with the scene significantly increases the time complexity. Our second main contribution addresses this through a novel approach, Simultaneous Multi-Agent Recurrent Trajectory (SMART) prediction. To the best of our knowledge, it is the first method to achieve multimodal, scene-consistent prediction of multiple agents in constant time.

Specifically, we propose a novel architecture based on Convolutional LSTMs (ConvLSTMs) \cite{convLSTM} and conditional variational autoencoders (CVAEs) \cite{CVAE}, where agent states and scene context are represented in the bird-eye-view. Our method predicts trajectories for $n$ agents with a time complexity of $O(1)$ (Table \ref{tab:time_complex}). To realize this, we use a single top-view grid map representation of all agents in the scene and utilize fully-convolutional operations to model the output predictions. Our ConvLSTM models the states of multiple agents, with novel state pooling operations, to implicitly account for interactions among objects and handle  dynamically changing scenarios. To obtain multimodal predictions, we assign labels to trajectories based on the type of maneuver they execute and query for trajectories executing specific behaviors at test time. Our variational generative model is conditioned on this label to capture diversity in executing maneuvers of various types.

We validate our ideas on both real and simulated datasets and demonstrate  state-of-the-art prediction numbers on both. We evaluate the network performance based on \emph{average displacement error}(ADE), \emph{final displacement error}(FDE) and \emph{likelihood}(NLL) of the predictions with respect to the ground truth. Our experiments are designed to highlight the importance of methods to simulate datasets with sufficient realism at larger scales and diversity, as well as a prediction method that accounts for multimodality while achieving constant-time outputs independent of the number of agents in the scene.

To summarize, our key contributions are:
\begin{itemize}
    \item A method to achieve constant-time trajectory prediction independent of number of agents in the scene, while accounting for multimodality and scene consistency.
    \item A method to simulate datasets in the top-view that imbibe the realism of real-world data, while augmenting them with diverse trajectories that cover diverse scene-consistent motion modes.
\end{itemize}

\begin{table}[t]
    \centering
    \tiny
    \caption{Comparison of our method with existing works in terms of complexity, scene context and interactions. n and K are number of agents and iterations.}
    \resizebox{0.98\textwidth}{!}
    {
    \begin{tabular}{c|c|c|c|c|c|c}
         \hline
         Method & Social GAN\cite{social_gan} & Desire\cite{DEISRE} & SoPhie\cite{sophiegan} & INFER\cite{INFER} & MATF GAN\cite{MATF} & Ours\\
         \hline
         Complexity & $O(n)$ & $O(nK)$ & $O(n^2)$ & $O(n)$ & $O(n)$ & $\boldsymbol{O(1)}$\\
         Scene Context & \xmark & \checkmark & \checkmark & \checkmark & \checkmark & \checkmark\\
         Social Interactions & \checkmark & \checkmark & \checkmark & \checkmark &  \checkmark & \checkmark\\
         \hline
    \end{tabular}
    }
    \label{tab:time_complex}
\end{table}

\section{Related Work}
% \buyu{We just copy the related work of our previous submission to here}
In this section, we briefly summarize datasets available for autonomous driving and talk about the existing forecasting techniques. 

\noindent{\bf Simulators and Autonomous Driving Datasets:} 
AirSim \cite{airSim} and CARLA \cite{carla} are autonomous driving platforms with primary target towards testing learning and control algorithms. SYNTHIA \cite{synthia} introduces a big volume of synthetic images with annotations for urban scenarios. Virtual KITTI \cite{VKITTI} imitates KITTI driving scenarios with varying environmental conditions and provide both pixel level and instance level annotations. Fang et al.\cite{Aug_lidar_AD} shows detectors trained with augmented lidar point cloud from a simulator provide comparable results with methods trained on real data. \cite{driving_in_matrix} generates images for vehicle detection and show an improvement in result that a deep neural network trained with synthetic data performs better than a network trained on real data when the dataset is bigger. Similarly, our focus is also to obtain better driving predictions by virtue of diversity from simulated datasets which emulate realistic driving behaviors. 

Until recently, KITTI \cite{KITTI} has been extensively used for evaluation of various computer vision applications like stereo, tracking and object detection, but has limited diverse behaviors for a scene. NGSIM \cite{colyar2007us} provides trajectory information of traffic participants but the scenes are only limited to highways with fixed lane traffic. CaliForecasting \cite{califorecasting} (unreleased) contains 10K examples with approximately 1.5 hours of driving data but does not contain any information about the scene. There are several recently proposed autonomous driving datasets \cite{rulesofroad_cvpr19,argoverse,traphic_cvpr19,nuscenes2019,ma2018trafficpredict,waymo_open_dataset,lyft2019}, some of which focus on trajectory forecasting \cite{rulesofroad_cvpr19,argoverse,traphic_cvpr19,ma2018trafficpredict}. Rules of the road \cite{rulesofroad_cvpr19} (unreleased) proposes a dataset with map information for approximately 83K trajectories in 88 distinct locations. Argoverse \cite{argoverse} proposes two datasets (Tracking and Forecasting) with HD semantic map information containing centerlines. Argoverse tracking contains a total of 113 scenes with tracklet information. While the forecasting dataset is sufficiently large enough with more than 300K trajectories, it contains 5 seconds trajectory data for only one vehicle in the scene. In our work, we simultaneously generate trajectories for all vehicles in the scene and provide trajectory information up to 7 seconds for each vehicle. NuScenes \cite{nuscenes2019} provides data from two cities with complete sensor suite information, but the main focus of the dataset is towards object detection and tracking. We primarily focus on using Argoverse Tracking\cite{argoverse} for simulating diverse trajectories and to showcase better prediction ability, but also simulate diverse trajectories for KITTI\cite{KITTI} dataset (please see supplementary material). Our method can be extended to many published datasets such as Waymo Open Dataset \cite{waymo_open_dataset} and Lyft \cite{lyft2019}.

\noindent{\bf Forecasting Methods:}
Motion forecasting has been extensively studied. Kitani et al.~\cite{kitani2012activity} proposes a method based on Inverse Optimal Control (IOC). Social LSTM \cite{Alahi_2016_CVPR} uses a recurrent network to model human-human interactions for pedestrian forecasting. Deo et al.~\cite{convSocialPooling} use a similar method as \cite{Alahi_2016_CVPR} to model interactions and predict an output distribution over future states for vehicles. DESIRE \cite{desire_cvpr17} uses a CVAE-based \cite{CVAE} approach to predict trajectories up to 4 seconds but requires multiple iterations to align its predictions with scene context. Sampling multiple trajectories that are semantically aligned might not be feasible.

Over the recent years, generative models \cite{social_gan,social_ways_cvprW19,sophiegan,socialbigat,which_way_cvpr19} have shown significant improvements in pedestrian trajectory prediction. Human trajectories tend to be stochastic and random while vehicle motions are aligned with the scene context and are strongly influenced by surrounding vehicle's behavior. Outputs from \cite{social_gan,sophiegan} show that it produces more diverse outputs and the output predictions are spread over a larger area. While more advanced methods \cite{socialbigat,trajectron} show outputs that are tightly coupled with the ground truth. We do not intend to capture the data distribution in such a fashion but are more focused towards producing predictions in possible dominant choices of motion\cite{sriramIV}. This also motivates us to use \cite{social_gan} to showcase the ability of simulation strategy in producing more diverse outputs. Our method capture both these indifference's by producing quintessential trajectories that are diverse and at the same time closely aligned with the ground truth (indicated by our likelihood values).

Recently, MATF GAN\cite{MATF} uses convolutions to model interactions between agents, but only shows results on straight driving scenarios and suffers in producing multimodal outputs. INFER \cite{INFER} proposes a method based on ConvLSTMs and \cite{rulesofroad_cvpr19} uses convolutions to regress future paths. These methods nicely couple scene context with predicted output, but their predictions are entity-centric and do not incorporate multi-agent stochastic predictions.

\begin{figure*}[t]
    \centering
    \includegraphics[width=1.0\textwidth]{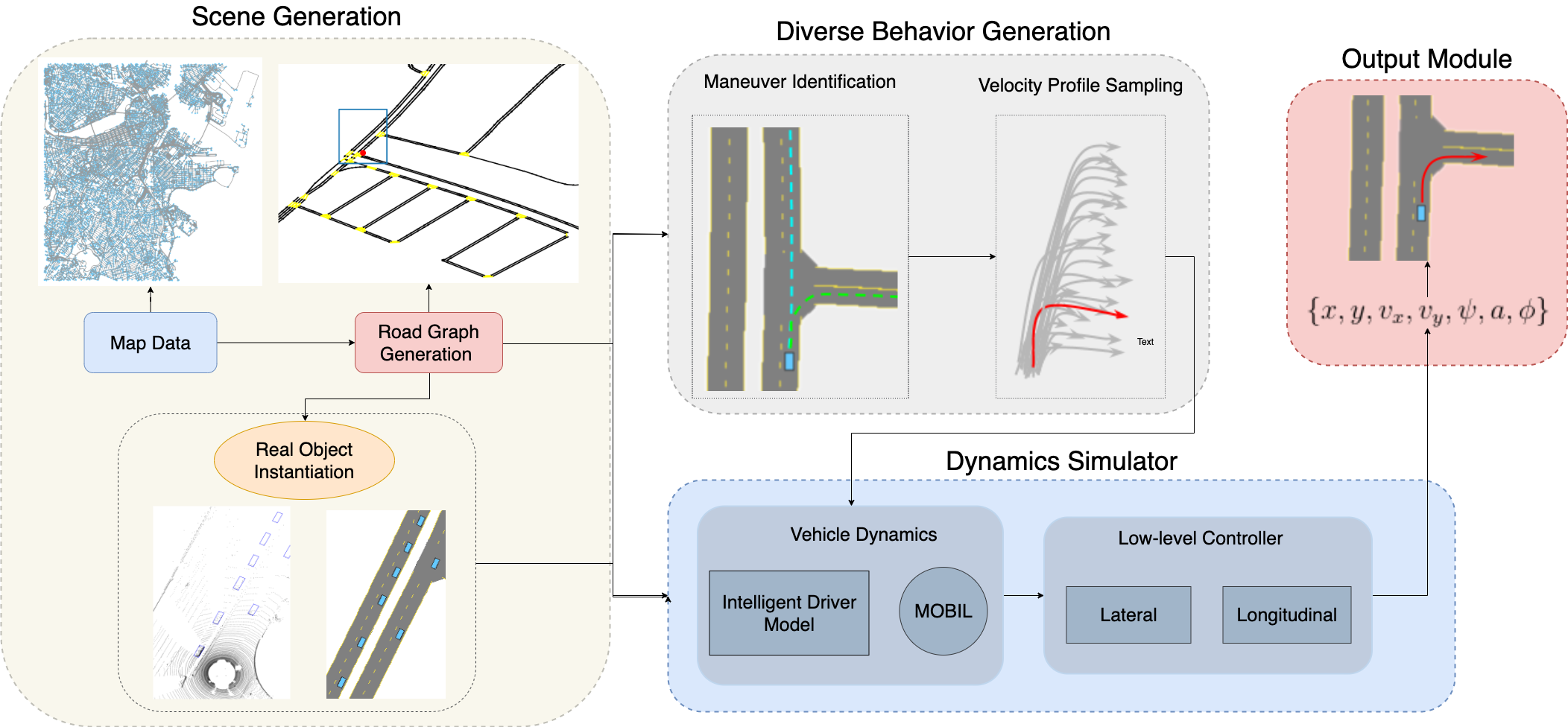}
    \caption{The overall pipeline of the proposed simulation strategy. Scene generation module recreates the specific driving scenarios from datasets. Behavior generation samples a behavior for simulation and an appropriate reference velocity profile for every vehicle in the scene. The dynamics simulator tracks the reference velocity and provides lane changing decisions based on the current traffic condition.}
    \label{fig:sim_pipeline}
\end{figure*}
\section{Simulated Dataset}

\begin{figure}
    \centering
    \begin{tabular}{cccccccc}
        \includegraphics[width=0.1\columnwidth, frame]{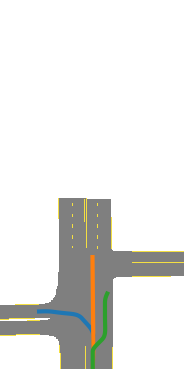} & 
        \includegraphics[width=0.1\columnwidth, frame]{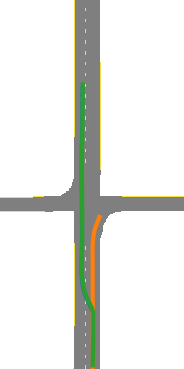} &
        \includegraphics[width=0.1\columnwidth, frame]{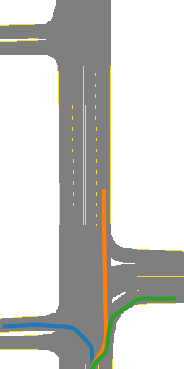} &
        \includegraphics[width=0.1\columnwidth, frame]{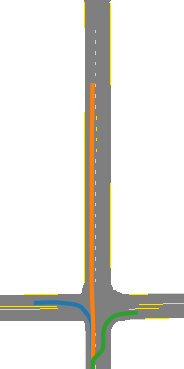} &
        \includegraphics[width=0.1\columnwidth, frame]{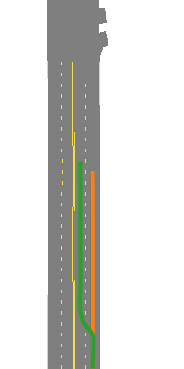} &
        \includegraphics[width=0.1\columnwidth, frame]{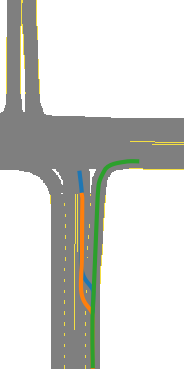} &
        \includegraphics[width=0.1\columnwidth, frame]{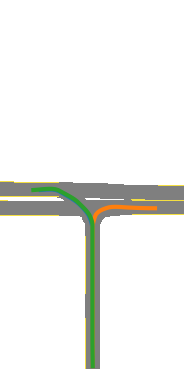}
    \end{tabular}
    \caption{Example trajectories executed by a single vehicle under different scenes in simulation.% The trajectories are shown from simulated vehicle's perspective for duration of 7 seconds.
    As shown in this figure, our simulation strategy is able to generate diverse yet realistic trajectories that align well with scene context.}
    \label{fig:traj_img}
\end{figure}

The overall pipeline of the proposed simulation strategy is shown in Figure \ref{fig:sim_pipeline}. Our simulation engine consists of three main components: scene generation module, behavior generation module and a dynamics simulation engine. Given a dataset to recreate and simulate, the scene generation module takes lane centerline information that can be either acquired through openly available map information\cite{OpenStreetMap} or provided by the dataset. We utilize this information to create a graph data structure that consists of nodes and edges representing end points of the lane and lane centerline respectively. This when rendered provides us with a Birds-Eye-View reconstruction of the local scene. We call this as {\it road graph}. The object instantiation module uses the tracklet's information from the dataset to project them on to the generated road graph. We do so by defining a coordinate system with respect to the ego vehicle and find the nearest edge occupied by the objects in our graph. This completes our scene reconstruction task. Now, for every vehicle that was instantiated in the scene, we find various possible maneuvers that it can execute given the traffic conditions and road structure from which, we uniformly sample different vehicle behaviors for our simulation. We refer to {\it behaviors} as vehicles executing different maneuvers like {\it straight, left turn, right turn and lane changes}. To execute such diverse behaviors that are significantly realistic, we sample appropriate velocity profiles from real dataset as references that closely resemble the intended behavior that vehicle is planning to execute. The dynamics simulation module utilizes this reference velocity to execute the right behavior for every vehicle but at the same time considers the scene layout and the current traffic conditions to provide a safe acceleration that can be executed. We simulate every scene for 7 seconds and generate a maximum of 3 diverse behaviors (Figure \ref{fig:traj_img}). The simulation is performed at 10Hz and output from our simulation consists of vehicle states $\{{\bf x},{\bf v},\psi,a,\phi\}_1^T$ which represent position, velocity, heading, acceleration and steering over the course of our simulation. We will now provide a brief description of each component and refer readers to supplementary material for additional details.

\vspace{-0.4cm}\subsubsection{Scene Generation}
We utilize the lane information from OpenStreetMaps (OSM) \cite{OpenStreetMap} or from datasets like \cite{argoverse} for creating the road graph. For our purposes, we make use of the road information such as centerline, number of lanes and one-way information for each road segment. Every bi-directional road centerline is split based on the specified number of lanes and one-way information. The vehicle pose information from the dataset is used to recreate exact driving scenarios. 
% We project them on to the local scene graph and use the tracklet's location and yaw information to find the nearest node that it occupies in the scene. For our simulation purposes, we only consider vehicles that are moving. These calculated assumptions make the scene reconstruction process practically feasible.

% \begin{figure}

% \begin{subfigure}{0.4\linewidth}
%     \includegraphics[width=\linewidth]{./images/left_right_vel.png}
%     \caption{}\label{fig:left_right_plot}
% \end{subfigure}
% \begin{subfigure}{0.2\linewidth}
%     \includegraphics[width=\linewidth]{./images/argo_distvel.png}
%     \caption{}\label{fig:dist_turn}
% \end{subfigure}
% \end{figure}

\vspace{-0.4cm}\subsubsection{Diverse Behavior Generation}
Given a particular lane ID (node) on the local road graph for every vehicle, we depth first explore $K$ possible leaf nodes that can be reached within a threshold distance. We categorize plausible maneuvers from any given node into three different categories \emph{\{left, right, straight\}}. Prior to the simulation, we create a pool of reference velocity profiles from the real data. At simulation time, after sampling a desired behavior, we obtain a Nearest Neighbor velocity profile for the current scene based on features such as distance before turn and average velocity, for turn and straight maneuvers respectively.

% \begin{figure}
%     \centering
%     \includegraphics[width=0.4\columnwidth, trim = 80mm 140mm 0mm 0mm, clip]{./images/traj_img.png}
%     \includegraphics[width=0.4\columnwidth, trim = 0mm 0mm 80mm 140mm, clip]{./images/traj_img.png}
%     \caption{Example trajectories executed by a single vehicle under different scenes in simulation.% The trajectories are shown from simulated vehicle's perspective for duration of 7 seconds.
%     As shown in this figure, our simulation strategy is able to generate diverse yet realistic trajectories that align well with scene context.}
%     \label{fig:traj_img}
% \end{figure}

\vspace{-0.4cm}\subsubsection{Dynamics Simulation}
The dynamics module utilizes road graph, a behavior from a pool of diverse plausible ones and a reference velocity that needs to be tracked for the appropriate behavior. Our dynamics engine is governed by Intelligent Driver Model (IDM)\cite{Treiber2000CongestedTS} and MOBIL\cite{mobil}. Acceleration and lane change decisions obtained from this dynamics module is fed to a low-level controller that tries to track and exhibit appropriate state changes in the vehicle behavior. In order to limit the acceleration under safety limit for the any traffic situation and to incorporate interactions among different agents in the scene we use an IDM\cite{Treiber2000CongestedTS} behavior for the simulated vehicles. 
% IDMs are Adaptive Cruise Control system which can adjust the driver's longitudinal velocity and safety time gap for safe and collision free maneuver. 
The input to an IDM consists of distance to the leading vehicle $s$, the actual velocity of the vehicle $v$, the velocity difference with the leading vehicle $\Delta v$ and provides an output $a_{IDM}$ that is considered safe for the given traffic conditions. It is given by the equation, 
\begin{equation}
    a_{IDM}(s,v,\Delta v) = a\Bigg(1 - \bigg(\frac{v}{v_o}\bigg)^{\delta} - \bigg(\frac{s^{*}(v,\Delta v)}{s}\bigg)^{2}\Bigg),
    \label{eq:a_idm}
\end{equation}
where, $a$ is the comfortable acceleration and $v_o$ is the desired reference velocity. $\delta$ is an exponent that influences how acceleration decreases with velocity. The deceleration of the vehicle depends on the ratio of desired minimum gap $s^{*}$ to actual bumper distance $s$ with the leading vehicle. 
% $s^{*}$ is given by the equation,
% \begin{equation}
%     s^{*}(v,\Delta v) = s_o + vT + \frac{v\Delta v}{2\sqrt{ab}},
%     \label{eq:s_star}
% \end{equation}
% where, $s_o$ is the desired safety distance to maintain, $T$ is the safety time gap of the vehicle and $b$ is the comfortable desired deceleration of the vehicle. $\{s_o, a, T, b, \delta\}$ are hyper-parameters that generate various vehicle behaviors. We sample these parameters during simulation time to generate simulations with different levels of aggressiveness.

{\it Lane Change Decisions:} We also consider lane changing behavior to add additional diversity in vehicle trajectories apart from turn based maneuver trajectories. Lane changing behaviors are modeled based on MOBIL algorithm from \cite{mobil}. The following are the parameters that control lane changing behavior: politeness factor $p$ that influences lane changing if there's acceleration gain for other agents, lane changing acceleration threshold $\Delta a_{th}$, maximum safe deceleration $b_{safe}$ and bias for particular lane $\Delta a_{bias}$. The following equations govern whether a lane change can be executed,

\begin{equation}
    \Tilde{a}_c - a_c + p\big\{ (\Tilde{a}_n - a_n) + (\Tilde{a}_o - a_o) \big\} > \Delta a_{th} - \Delta a_{bias},
\end{equation}
\begin{equation}
    (\Tilde{a}_n - a_n) > -b^n_{safe}, (\Tilde{a}_c - a_c) > -b^c_{safe}.
\end{equation}
Here, $a$ is the current acceleration and $\Tilde{a}$ represents the new acceleration after lane change. $c,n,o$ subscripts denote current, new vehicle and old vehicles respectively.

\begin{figure*}[t]
    \centering
    \includegraphics[width=1.0\textwidth]{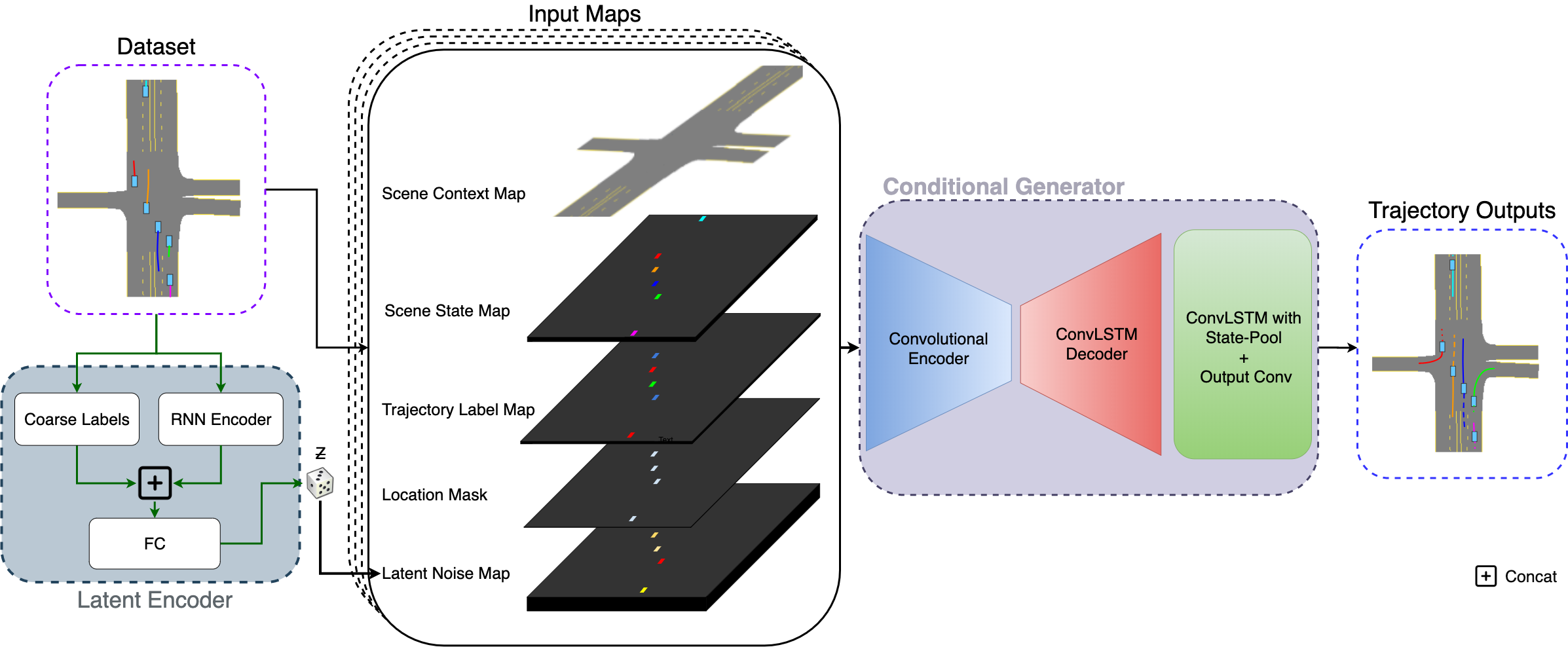}
    \caption{The overall architecture for SMART framework. The components connected in green are used only during the training phase. It takes in a single representation of the scene, to regress per timestep coordinates for the all agents at their respective location in the spatial grid.}
    \label{fig:smart_pipeline}
\end{figure*}
\section{SMART}

In this section, we will introduce a single representation model to predict trajectories for {\it multiple} agents in a road scene such that our predictions are {\it context aware}, {\it multimodal} and have {\it constant inference time} irrespective of number of agents. We formulate the trajectory prediction problem as per frame regression of agents locations over the spatial grid. We will describe our method below in details.%We now detail our formulation and explain our architecture broadly.

\subsection{Problem formulation}

Given the lane centerline information $L^{1...m}$ for a scene, we render them in top view representations such that our scene context map $\mathcal{I}$ is of HxWx3 where channel dimension represents one-hot information of each pixel corresponding to \{{\it road}, {\it lane},{\it unknown}\} road element. Let ${\bf X}_i = \{X_i^1,X_i^2,...,X_i^T\}$ denote trajectory information of $i^{th}$ vehicle from timestep $1...T$ where each $X_i^t=(x_i,y_i)^t$ represents spatial location of the agent in the scene. Our network takes input in the form of relative coordinates $^R{\bf X}_i$ with respect to agent's starting location. 
For the $i^{th}$ agent in the scene, we project $^R{\bf X}_i$ at corresponding ${\bf X}_i$ locations to construct a spatial location map of states $\mathcal{S}^{1...T}$ such that $\mathcal{S}^t[{X}_i^t]$ contains relative coordinate of $i^{th}$ agent at timestep $t$. $^R{\bf Y}_i = ^R{\bf X}_i^{t_{obs}...T}$ represents ground truth trajectory. And we further denote $\mathcal{M}^t$ as the location mask representing configuration of agents in the scene. To keep track of vehicles across timesteps, we construct a vehicle IDs map $\mathcal{V}^{1...T}$ where $\mathcal{V}^t[X_i^t] = i$. Furthermore, we associate each trajectory $X_i^{t_{obs},..T}$ with a label $c_i$ that represents the behavioral type of the trajectory from one of \{{\it straight, left, right}\} behaviors. And trajectory label for lane changes falls in one of the three categories. Let $\mathcal{C}$ encode grid map representation of $c_i$ such that $\mathcal{C}^t[{X}_i^t] = c_i$. Note that vehicle trajectories are not random compared to the human motion. Instead, they depend on behaviors of other vehicles in the road, which motivates us to classify trajectories based on different maneuvers.
%On a macroscopic level these trajectories are similar and depend on the road structure and traffic conditions.

%We formulate the problem similar to previous works
We follow the formulation proposed in~\cite{MATF,rulesofroad_cvpr19,INFER} where network takes previous states $\mathcal{S}^{1..t_{obs}}$ as input along with the scene context map $\mathcal{I}$, trajectory label map $\mathcal{C}$, location mask $\mathcal{M}$ and a noise map $\mathcal{Z}$ to predict the future trajectories $^R\hat{{\bf Y}}_i$ for every agent at its corresponding grid map location $X_i^t$ in the scene. Note that we do not have a separate head for each agent. Instead, our network predicts a single future state map $\hat{\mathcal{S}}^t$ where each individual agent tries to match $^R{\bf Y}_i^t$ at $t$. 

%We're interested in predicting trajectories of agents in the road scenes in a simultaneous fashion for which create a single joint representation of the complete scene. 
% In order to predict trajectories of multiple agents simultaneously w.r.t. road structure, we first project trajectory coordinates in the spatial grid. Specifically, we embed the relative coordinates in the spatial grid, which enables us to discretize the locations of the agents in the scene but also keep the original information in the mean time.~\buyu{Not so sure whether this is true or not.} 
% %do not lose information due to discretization as we embed relative coordinates in the spatial grid. 
% Our representation provides a sequence of scene state map where inside each timestep all agents in the scene are presented. This joint representation is compact yet powerful as it provides direct correspondence of the agents location with the scene context.~\buyu{We should finish introducing our representation before the method part.} 

\subsection{Method}
% \sriram{Very roughly written}

% Vehicle forecasting is a challenging and ambiguous problem to tackle as vehicle motions are goal oriented and can have multiple plausible maneuvers at various situations. Hence, capturing multimodality is an essential 

We illustrate our pipeline in Figure \ref{fig:smart_pipeline}. 
% Previous method\cite{INFER} uses a similar representation to capture the scene context in concurrence with agent's state but is incapable of multiple agent prediction in one shot. 
Our network architecture comprises of two major parts, a latent encoder and a conditional generator. We model the temporal information with the agents previous locations using ConvLSTMs. We further introduce a state pooling operation to feed agents state information at respective locations in consecutive timestep.
%The agents location in the scene varies with every timestep and as there are multiple agents in the scene we introduce a state pooling operation to feed agents state information back at respective locations in the next consecutive timestep. 
While we provide trajectory specific labels to capture diverse predictions,
%on a macroscopic level, 
we leverage conditional variational generative models (CVAE\cite{CVAE}) to model diversity in the data for each type of label.

\vspace{-0.4cm}\subsubsection{Latent Encoder:} It acts as a recognition module $Q_{\phi}$ for our CVAE framework and is only used during our training phase. Specifically, it takes in both the past and future trajectory information $^R{\bf X}_i$ and passes them through an embedding layer. %to encode them to higher dimensions. 
The embedded vectors are then passed on to a LSTM network to output encoding at every timestep. The outputs across all the timesteps are concatenated together into a single vector along with the one hot trajectory label $c_i$ to produce $V_{enc}(i)$. This vector is then passed on through a MLP to obtain $\mu$ and $\sigma$ to output a distribution $Q_{\phi}(z_i\vert ^R{\bf X}_i, c_i)$. Formally,

\begin{equation}
\begin{gathered}
    ^{o}h_i^t = LSTM(h_i^{t-1}, ^RX_i^t)\\
    V_{enc}(i) = [^oh_i^1,...,^oh_i^T, c_i]\\
    \mu, \sigma = MLP(V_{enc}(i)).
\end{gathered}
\end{equation}

\vspace{-0.4cm}\subsubsection{\bf Conditional Generator:} 
We adapt a U-Net like architecture for the generator. At any timestep $t$, the inputs to the network conditional generator are the following, a scene context map $\mathcal{I}$ (HxWx3), a single representation of all agents current state $\mathcal{S}^t$ (HxWx2), location mask $\mathcal{M}^t$ (HxWx1), a one-hot trajectory specific label for each agent projected at agent specific locations in a grid from $\mathcal{C}^t$ (HxWx3) and a latent vector map $\mathcal{Z}^t$ (HxWx16) containing $z_i$ obtained from $Q_{\phi}(z_i\vert ^R{\bf X}_i, c_i)$ during training phase or sampled from prior distribution $P_{v}(z_i \vert ^R{\bf X}_i, c_i)$ at test time. Formally the network input $E^t$ is given by:
\begin{equation}
    E^t = [\mathcal{I}, \mathcal{S}^t, \mathcal{M}^t, \mathcal{C}^t, \mathcal{Z}^t],
\end{equation}
which is of size HxWx25 for any timestep $t$. Note that our representation is not entity centric i.e we do not have one target entity for which we want to predict trajectories but rather have a global one for all agents. %Instead, our maps are centered with the ego vehicles location to enable predictions around the vehicle of interest. 
% We leverage convolutional operations to model interactions of agents with scene and other agents. 

At each timestep from $1,...,t_{obs}$, we pass the above inputs through the encoder module. %Our encoder module has alternative strided convolutions that downsizes the input to encode information into a low dimensional feature map. 
This module is composed of strided convolutions, which encode information in small spatial dimensions, and passes them through the decoder. 
The decoder includes ConvLSTMs and transposed convolutions with skip connections from the encoder module, and outputs a HxW map.
%The output size of the decoder has same spatial dimension HxW as that of the input. 
It is then passed on to another ConvLSTM layer with state pooling operations. The same network is shared during observation and prediction phase. A final 1x1 convolution layer is added to output a 2 channel map containing relative predicted coordinates $^RX_i^t$ for the agents in the next timestep. 

We use the ground truth agent locations for the observed trajectory and unroll our ConvLSTM based on the predictions of our network. During the prediction phase ($t_{obs},...,T$), the outputs are not directly fed back as inputs to the network rather the agent's state is updated to the next location in the scene based on the predictions. The relative predicted location $^R{\hat{X}}_i^{t-1}$ gets updated to absolute predicted location ${\hat{X}}_i^{t}$ to obtain a updated scene state map $\hat{\mathcal{S}}^t$ containing updated locations of all the agents in the scene. Note that using such representations for the scene is agnostic to number of agents and as the agents next state is predicted at its respective pixel location it is capable of handling dynamic entry and exit of agents from the scene. 

\vspace{-0.4cm}\subsubsection{State-Pooled ConvLSTMs:} Simultaneous multi-agent predictions are realized through state-pooling in ConvLSTMs. Using standard ConvLSTMs for multi-agent trajectory predictions usually produces semantically aligned trajectories, but the trajectories occasionally contain erratic maneuvers. We solve this issue via state-pooling, which ensures the availability of previous state information when trying to predict the next location. We pool the previous state information from the final ConvLSTM layer for all the agents $^{sp}{\bf H}_i^{t-1}$ and initialize the next state with $^{sp}{\bf H}_i^{t-1}$ (for both hidden and cell state) at agents updated locations and zero vectors at all other locations for timestep t.

\vspace{-0.4cm}\subsubsection{Learning:} We train both the recognition network $Q_{\phi}(z_i\vert ^R{\bf X}_i, c_i)$ and the conditional generator $P_{\theta}(Y|E)$ concurrently. We obtain predicted trajectory $^R\hat{{\bf Y}}$ by pooling values from indexes that agents visited at every timestep. We use two loss functions in training our CVAE based ConvLSTM network:
\begin{itemize}
    \item Reconstruction Loss: $\mathcal{L}_R = \frac{1}{N}\sum_i^N|| ^R{\bf Y}_i - ^R\hat{{\bf Y}}_i||$ that penalizes the predictions to enable them to reconstruct the ground truth accurately.
    \item KL Divergence Loss: $\mathcal{L}_{KLD} = D_{KL}(Q_{\phi}(z_i\vert ^R{\bf X}_i, c_i)||P_{v}(z_i | ^R{\bf X}_i, c_i))$ . That regularizes the output distribution from $Q_{\phi}$ to match the sampling distribution $P_v$ at test time. 
\end{itemize}

\vspace{-0.4cm}\subsubsection{Test phase:} At inference time, we do not have access to trajectory specific labels $c_i$ but rather query for a specific behavior by sampling these labels randomly. Along with $c_i$ for each agent we also sample $z_i$ from $P_v(z_i | ^R{\bf X}_i, c_i)$. However, $P_v$ can be relaxed to be independent of the input\cite{CVAE} implying the prior distribution to be $P_v(z_i)$. $P_v(z_i):=\mathcal{N}(0,1)$ at test time. 

% Unlike existing works\cite{INFER,MATF,social_gan,DEISRE}, we do not directly feed the output predictions back to the input, but update the agents locations and its predictions to next location in the spatial grid containing $^R{\hat{X}}_i^{t-1}$ at ${\hat{X}}_i^{t}$ to obtain a next state map $\hat{\mathcal{S}}^t$ containing all the agents at their updated locations in the scene.~\buyu{What do you mean here?} 

% The state pooling is necessary to provide agents an idea of how agents state changes over time. Formally, 
% \begin{equation}
%     ^{s}h
% \end{equation}

% The output state from this layers is passed through a final 1x1 convolution to output a 2 channel map containing relative predicted coordinates for the agents in the next timestep. The 
% \sriram{Explain the flow. Introduce the loss functions and explain state pooling in equations.}

\section{Experiments}

\begin{table}[t]
\caption{Quantitative measurements on P-ArgoT. We report ADE, FDE (in meters) and NLL (N=5)}
\tiny
\centering
\resizebox{1.0\textwidth}{!}
{
\begin{tabular}{c ||c|c|c|| c|c|c|| c|c|c|| c|c|c ||c|c|c}
\hline
% \tiny
{Model} & \multicolumn{3}{c||}{1.0(sec)} & \multicolumn{3}{c||}{2.0(sec)} & \multicolumn{3}{c||}{3.0(sec)} & \multicolumn{3}{c||}{4.0(sec)} & \multicolumn{3}{c}{5.0(sec)}\\ 
    \hline
    \multicolumn{16}{c}{P-ArgoT $\|$ ADE $\vert$ FDE $\vert$ NLL$\|$ }\\
    \hline
    LSTM & 0.53 & 0.87 & - & 1.03 & 2.01 & - & 1.62 & 3.41 & - & 2.31 & 5.09 & - & 3.09 & 6.98 & -\\
    CVAE & 0.46 & 0.73 & {\bf 2.16} & 0.89 & 1.72 & {\bf 3.31} & 1.42 & 2.98 & 4.18 & 2.04 & 4.49 & 4.88 & 2.76 & 6.26 & 5.48\\
    MATF Scene \cite{MATF} & 0.98 & 1.73 & - & 1.84 & 3.53 & - & 2.76 & 5.48 & - & 3.72 & 7.56 & - & 4.73 & 9.78 & -\\
    MATF GAN \cite{MATF}& 0.78 & 1.34 & 3.44 & 1.45 & 2.73 & 4.53 & 2.17 & 4.28 & 5.24 & 2.94 & 5.95 & 5.79 & 3.76 & 7.77 & 6.23\\
    S-GAN\cite{social_gan} & {\bf 0.42} & 0.72 & 2.21 & 0.85 & 1.68 & 3.49 & 1.36 & 2.83 & 4.36 & 1.93 & 4.08 & 5.03 & 2.54 & 5.46 & 5.57\\
    \hline 
    SMART $(c_{random})$ & 0.73 & 0.64 & 3.72 & 0.84 & 0.98 & 3.92 & 0.94 & 1.35 & 4.31 & 1.15 & 1.73 & 4.70 & 1.38 & 2.16 & 5.06\\
    SMART $(c_{best})$ & 0.58 & {\bf 0.59} & 3.21 & {\bf 0.59} & {\bf 0.55} & 3.39 & {\bf 0.60} & {\bf 0.75} & {\bf 3.63} & {\bf 0.98} & {\bf 0.61} & {\bf 3.89} & {\bf 1.02} & {\bf 1.06} & {\bf 4.13}\\
    \hline
\end{tabular}
}
\label{tab:P-argoT}
\end{table}

We evaluate our methods on publicly available Argoverse\cite{argoverse} Tracking(ArgoT)~\footnote{Generated 2044 scenes in total containing multiple trajectories for every scene} and Forecasting(ArgoF)~\footnote{Argoverse Forecasting for vehicle trajectory prediction is a large scale dataset containing 333,441 (5sec) trajectories captured from 320 hours of driving.} dataset.% a simulation validation set generated using our simulation strategy. 
%We simulate for a total of 
We also introduce a simulated dataset based on P-ArgoT and conduct experiments with it. Our simulated dataset utilizes 2000 scene instances from ArgoT to generate scenarios with multiple agents and trajectory durations of 7 seconds.%
%We utilize Argoverse Tracking\cite{argoverse} to show 
% We evaluate the simultaneous multi-agent prediction capabilities of our SMART algorithm on ArgoT and further demonstrate that with the help from simulated data, we can achieve high quality yet more diverse predictions.
% We also show that having diversity in the data can help in providing better prediction results but at the same time is also contingent on the methods ability in capturing such diversity in the right fashion. 

We use standard {\it evaluation metrics} suggested in previous approaches\cite{argoverse,MATF,INFER,social_gan},%The metrics are
e.g. Average Displacement Error(ADE), Final Displacement Error(FDE) and Negative Log Likelihood(NLL) with the ground truth.
% {\bf Metrics for evaluation:} We use standard evaluation metrics used in previous approaches\cite{argoverse,MATF,INFER,social_gan} to evaluate the efficacy of our method and to show performance gain obtained by training our method on simulated data. The metrics are,
% \begin{itemize}
%     \item {\it Average Displacement Error (ADE):} Mean L2 distance our predictions with that of the ground truth.
%     \item {\it Final Displacement Error (FDE):} L2 distance between final location of the predicted trajectories and that of the ground truth.
%     \item {\it Negative Log Likelihood (NLL):} The average log likelihood of the ground truth with that of the predicted trajectory calculated across  prediction timesteps. 
% \end{itemize}

We evaluate two versions of {\it SMART}, e.g. (SMART($c_{random}$)) and (SMART($c_{best}$)). For the former, we randomly sample our behavior specific trajectory labels for evaluation, while for the later we equally sample $n$ trajectories over all the trajectory labels and report the best results across all. We comapare our proposed methods against the following {\it baselines}: %We test our generated dataset and method with the following baselines:
\begin{itemize}
    \item LSTM: A sequence to sequence encoder-decoder network that regresses future locations based on the past trajectory \cite{lstm_seq}.
    \item CVAE: A modified LSTM generator that predicts paths based on the input latent vector in the form of noise learned from the data distribution \cite{CVAE}.
    \item S-GAN\cite{social_gan}: We implement and evaluate this method on all datasets.
    %The method proposed in \cite{social_gan}.%A human trajectory prediction method that incorporates hidden states of other agents in the scene using a pooling module. A GAN based method that uses noise vectors fed into the initial state to generate future trajectories. 
    
    \item MATF GAN\cite{MATF}: We implement it ourselves and evaluate it on all datasets.
    %The method proposed in \cite{MATF}.%A state-of-the-art trajectory prediction method proposed for human and vehicle trajectory prediction. Incorporates multi-agent information using convolutional networks. 
\end{itemize}

\begin{table}[t!]
\centering
\caption{Left: Quantitative measurements on ArgoF validation set. (N=6). Right: Quantitative comparison of different datasets with introduced diversity metrics based on wasserstein distances.}
\scalebox{0.75}{
% \begin{tabular}{c ||c|c|c|| c|c|c|| c|c|c}
% \hline
% {Model} & \multicolumn{3}{c||}{1.0(sec)} & \multicolumn{3}{c||}{2.0(sec)} & \multicolumn{3}{c}{3.0(sec)}\\ 
%     \hline
%     \multicolumn{10}{c}{Argo Forecasting Dataset (ArgoF) \|ADE \vert FDE \vert NLL\| }\\
%     \hline
%     LSTM & 0.67 & 1.09 & - & 1.24 & 2.49 & - & 2.00 & 4.45 & -\\
%     CVAE & 1.05 & 1.91 & 10.3 & 2.16 & 4.47 & 12.3 & 3.50 & 7.61 & 13.6\\
%     MATF Scene \cite{MATF} & 1.56 & 2.71 & - & 2.90 & 5.54 & - & 4.35 & 8.17 & -\\
%     MATF GAN \cite{MATF} & 1.48 & 2.54 & 13.5 & 2.72 & 5.17 & 13.6 & 4.08 & 8.13 & 14.0\\
%     S-GAN \cite{social_gan} & {\bf 0.51} & {\bf 0.72} & {\bf 3.10} & 1.01 & 1.91 & 4.36 & 1.70 & 3.57 & 5.23\\
%     \hline
%     SMART (c_{random}) & 0.77 & 0.88 & 3.64 & 1.10 & 1.65 & 4.71 & 1.56 & 2.72 & 5.29\\
%     SMART (c_{best}) & 0.71 & 0.75 & 3.19 & {\bf 0.99} & {\bf 1.39} & {\bf 3.60} & {\bf 1.39} & {\bf 2.32} & {\bf 4.07}\\
%     \hline
% \end{tabular}
\begin{tabular}{c ||c|c|c|| c|c|c|| c|c|c}
\hline
{Model} & \multicolumn{3}{c||}{1.0(sec)} & \multicolumn{3}{c||}{2.0(sec)} & \multicolumn{3}{c}{3.0(sec)}\\ 
    \hline
    \multicolumn{10}{c}{Argo Forecasting Dataset (ArgoF) $\|$ ADE $\vert$ FDE $\vert$ NLL$\|$ }\\
    \hline
    LSTM & 0.76 & 1.16 & - & 1.32 & 2.67 & - & 2.14 & 4.71 & -\\
    CVAE & 1.22 & 2.27 & 9.14 & 2.56 & 5.32 & 11.7 & 4.14 & 8.94 & 13.2\\
    MATF Scene \cite{MATF} & 1.56 & 2.71 & - & 2.90 & 5.54 & - & 4.35 & 8.17 & -\\
    MATF GAN \cite{MATF} & 1.48 & 2.54 & 13.5 & 2.72 & 5.17 & 13.6 & 4.08 & 8.13 & 14.0\\
    S-GAN \cite{social_gan} & 0.88 & 1.59 & 4.12 & 1.99 & 4.34 & 5.74 & 3.49 & 8.05 & 6.80\\
    \hline
    % SMART (c_{random}) & 0.62 & 0.69 & 4.18 & 0.86 & 1.35 & 4.14 & 1.27 & 2.54 & 4.56\\
    % SMART (c_{best}) & {\bf 0.55} & {\bf 0.57} & {\bf 3.24} & {\bf 0.72} & {\bf 1.05} & {\bf 3.60} & {\bf 1.05} & {\bf 1.98} & {\bf 4.16}\\
    SMART $(c_{random})$ & 0.79 & 0.96 & 4.59 & 1.16 & 1.85 & 4.76 & 1.65 & 3.00 & 5.18\\
    SMART $(c_{best})$ & {\bf 0.71} & {\bf 0.83} & {\bf 3.56} & {\bf 1.03} & {\bf 1.55} & {\bf 4.05} & {\bf 1.44} & {\bf 2.47} & {\bf 4.61}\\
    \hline
\end{tabular}

\begin{tabular}{c|cc|cc}
\hline
Datasets & \multicolumn{2}{c|}{Y Wasserstein} & \multicolumn{2}{c|}{$\Ddot{X}$ Wasserstein}\\
\hline
& Mean & Median & Mean & Median\\
\hline
KITTI & 0.14 & 0.04 & 4.91 & 3.52\\
P-KITTI & {\bf 2.13} & {\bf 0.75} & {\bf 17.64} & {\bf 17.58}\\
\hline
ArgoT & 0.49 & {\bf 0.20} & 5.98 & 2.97\\
P-ArgoT& {\bf 0.97} & 0.12 & {\bf 17.5} & {\bf 17.49}\\
\hline
% Argo Forecasting & 0.60 & 0.14 & {\bf 35.28} & {\bf 29.43}\\
% P-Argo Forecasting & {\bf 1.26} & {\bf 0.23} & 17.53 & 17.45\\
% \hline    
\end{tabular}}
\label{tab:argoF}
\end{table}

\begin{table}[h]
\caption{Results for methods tested on ArgoT. `[ ]' represents the training set. We report results on the basis of ADE, FDE and NLL (N=5).}
\centering
% \tiny
\resizebox{0.98\textwidth}{!}{
\begin{tabular}{c ||c|c|c|| c|c|c|| c|c|c|| c|c|c ||c|c|c}
\hline
{Model} & \multicolumn{3}{c||}{1.0(sec)} & \multicolumn{3}{c||}{2.0(sec)} & \multicolumn{3}{c||}{3.0(sec)} & \multicolumn{3}{c||}{4.0(sec)} & \multicolumn{3}{c}{5.0(sec)}\\ 
    \hline
    \multicolumn{16}{c}{ArgoT $\|$ ADE $\vert$ FDE $\vert$ NLL $\|$ }\\
    \hline
    LSTM $[$ArgoT$]$ & 0.65 & 1.07 & - & 1.28 & 2.53 & - & 2.07 & 4.45 & - & 3.00 & 6.74 & - & 4.05 & 9.31 & -\\
    CVAE $[$ArgoT$]$ & {\bf 0.45} & 0.75 & {\bf 1.99} & 0.90 & 1.88 & {\bf 3.13} & 1.52 & 3.48 & {\bf 4.07} & 2.30 & 5.50 & 4.89 & 3.21 & 7.89 & 5.63\\
    MATF Scene $[$ArgoT$]$ \cite{MATF} & 1.24 & 2.20 & - & 2.39 & 4.67 & - & 3.66 & 7.49 & - & 5.03 & 10.4 & - & 6.45 & 13.4 & -\\
    \hline
    MATF GAN $[$ArgoT$]$ \cite{MATF} & 0.97 & 1.69 & 5.32 & 1.82 & 3.47 & 6.21 & 2.75 & 5.53 & 6.93 & 3.77 & 7.88 & 7.58 & 4.90 & 10.5 & 8.15\\
    MATF GAN $[$P-ArgoT$]$ \cite{MATF} & 1.03 & 1.79 & 6.32 & 1.93 & 3.68 & 7.51 & 2.89 & 5.77 & 8.43 & 3.94 & 8.13 & 9.19 & 5.07 & 10.7 & 9.77\\
    \hline
    S-GAN $[$ArgoT$]$ \cite{social_gan} & 0.77 & 1.35 & 4.29 & 1.47 & 2.79 & 5.48 & 2.25 & 4.47 & 6.24 & 3.11 & 6.38 & 6.81 & 4.06 & 8.54 & 7.27\\
    S-GAN $[$P-ArgoT$]$ \cite{social_gan} & 0.94 & 1.63 & 4.84 & 1.76 & 3.31 & 6.00 & 2.66 & 5.24 & 6.74 & 3.66 & 7.37 & 7.30 & 4.74 & 9.68 & 7.75\\
    \hline
    SMART $[$ArgoT$]$ $(c_{random})$ & 0.85 & 1.06 & 4.31 & 1.22 & 1.87 & 4.82 & 1.68 & 2.98 & 5.38 & 2.25 & 4.30 & 5.88 & 2.88 & 5.70 & 6.31\\
    SMART $[$P-ArgoT$]$ $(c_{random})$ & 0.68 & 0.80 & 4.12 & 0.97 & 1.51 & 4.29 & 1.37 & 2.45 & 4.71 & 1.85 & 3.58 & 5.13 & 2.39 & 4.85 & 5.51\\
    \hline
    SMART $[$ArgoT$]$ $(c_{best})$ & 0.74 & 0.87 & 3.85 & 1.03 & 1.50 & 4.02 & 1.42 & 2.40 & 4.42 & 1.90 & 3.52 & 4.84 & 2.45 & 4.74 & 5.24\\
    SMART $[$P-ArgoT$]$ $(c_{best})$ & 0.61 & {\bf 0.66} & 3.91 & {\bf 0.84} & {\bf 1.21} & 3.81 & {\bf 1.16} & {\bf 1.97} & 4.08 & {\bf 1.56} & {\bf 2.91} & {\bf 4.39} & {\bf 2.02} & {\bf 3.96} & {\bf 4.70}\\
    \hline

\end{tabular}
}
\label{tab:argoT}
\end{table}

\begin{table}[h]
\caption{Results for methods tested on ArgoT without straight trajectories. `[ ]' represents the training set. We report results on the basis of ADE, FDE(N=5).}
\centering
% \tiny
% \tiny
\resizebox{0.85\textwidth}{!}{
\begin{tabular}{c ||c|c|| c|c|| c|c|| c|c ||c|c}
\hline
{Model} & \multicolumn{2}{c||}{1.0(sec)} & \multicolumn{2}{c||}{2.0(sec)} & \multicolumn{2}{c||}{3.0(sec)} & \multicolumn{2}{c||}{4.0(sec)} & \multicolumn{2}{c}{5.0(sec)}\\ 
    \hline
    \multicolumn{11}{c}{ArgoT $\|$ ADE $\vert$ FDE $\|$ }\\
    \hline
    MATF GAN [ArgoT] \cite{MATF} & 1.04 & 1.80 & 1.98 & 3.86 & 3.08 & 6.50 & 4.41 & 9.91 & 5.98 & 14.1\\
    MATF GAN [P-ArgoT] \cite{MATF} & 0.94 & 1.63 & 1.79 & 3.50 & 2.81 & 6.00 & 4.07 & 9.35 & 5.61 & 13.6\\
    \hline
    S-GAN [ArgoT] \cite{social_gan} & 0.94 & 1.67 & 1.86 & 3.64 & 2.91 & 5.91 & 4.12 & 8.48 & 5.47 & 11.4\\
    S-GAN [P-ArgoT]\cite{social_gan} & 0.93 & 1.61 & 1.74 & 3.30 & 2.65 & 5.23 & 3.65 & 7.38 & 4.73 & 9.75\\
    \hline
\end{tabular}
}
\label{tab:argoT_no_straight}
\end{table}

\begin{table}[b!]
    \centering
    % \tiny
    \caption{Average runtime in seconds to generate one prediction sample in scenes from Argoverse\cite{argoverse} dataset with increasing number of agents, benchmarked on RTX2080Ti, 11GB GPU.}
    \begin{tabular}{c  @{\hspace{0.3cm}} |  @{\hspace{0.3cm}} c @{\hspace{0.3cm}} c @{\hspace{0.3cm}} c @{\hspace{0.3cm}} c @{\hspace{0.3cm}} c @{\hspace{0.3cm}} c @{\hspace{0.3cm}} c @{\hspace{0.3cm}} c @{\hspace{0.3cm}} c @{\hspace{0.3cm}} c}
         No. of Agents & 1 & 2 & 3 & 4 & 5 & 6 & 7 & 8 & 9 & 10\\
    \hline
         SMART & .070 & .070 & .070 & .070 & .072 & .069 & .069 & .075 & .072 & .069\\  
         S-GAN\cite{social_gan} & .024 & .034 & .044 & .054 & .062 & .071 & .082 & .090 & .102 & .108\\
         %MATF GAN\cite{MATF} & .014 & .014 & .015 & .015 & .015 & .015 & .016 & .016 & .016 & .016\\
    \end{tabular}
    \label{tab:mean_runtime}
\end{table}

\subsubsection{Quantitative Results: }
\begin{figure}[b!]
    \centering
    \begin{tabular}{cc}
        \includegraphics[width=0.5\linewidth, height=0.3\linewidth]{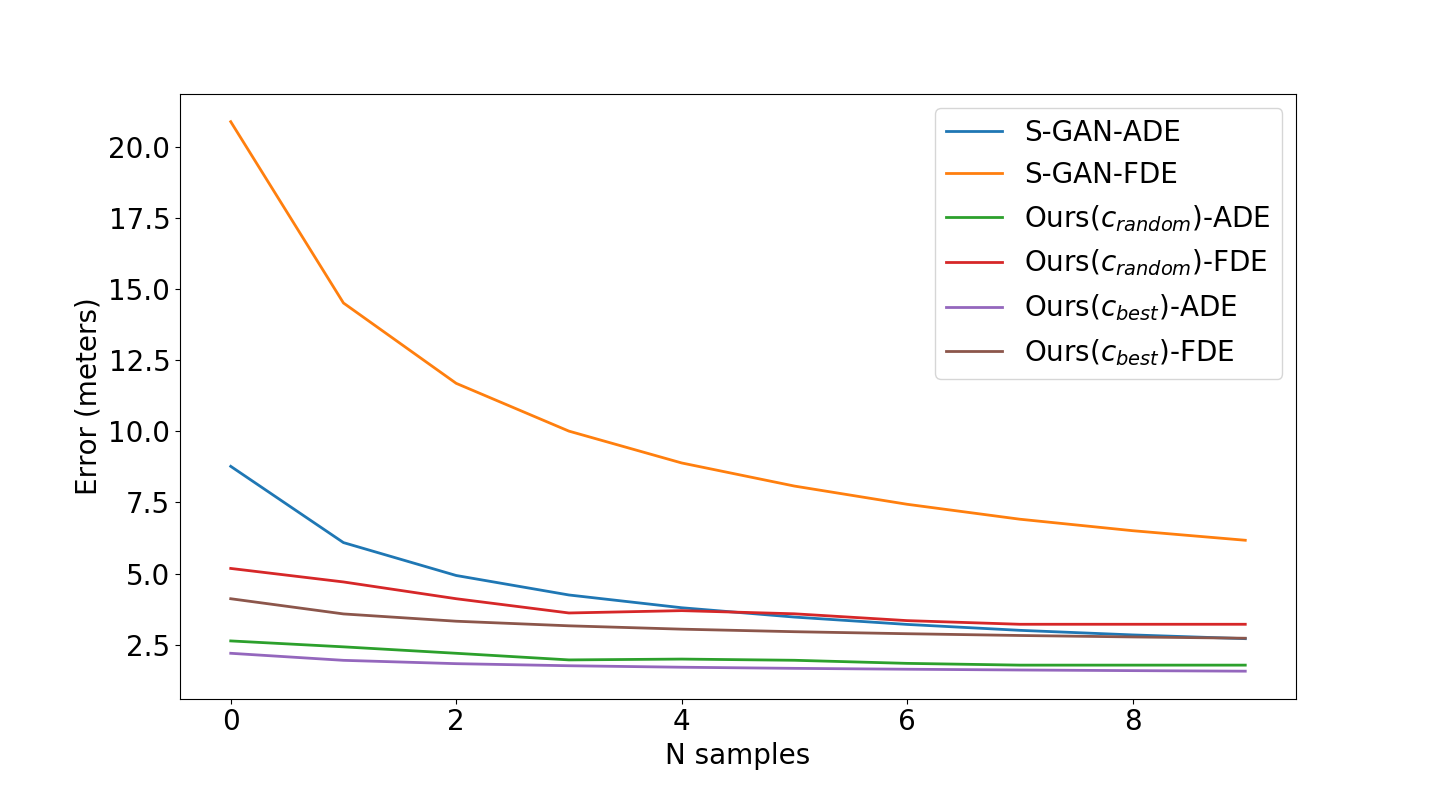} & 
        \includegraphics[width=0.5\linewidth, height=0.3\linewidth]{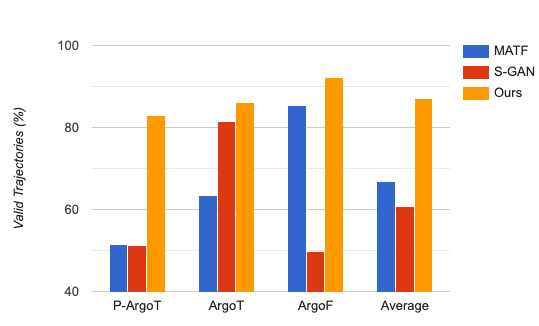}\\
    \end{tabular}
    \caption{{\bf Left:}Quantitative results on ArgoF with increasing number of samples. Average and final displacements of our method is plotted against S-GAN\cite{social_gan}. {\bf Right:} Percentage of samples(n=30) that produced trajectories inside the road.}
    \label{fig:ade_valid}
\end{figure}

We first demonstrate that our proposed method,  {\bf SMART}, can beat the baseline methods. As shown in Tab.~\ref{tab:P-argoT} and~\ref{tab:argoF}(left), %compare prediction numbers of the method with the baselines for both real and simulated datasets. 
our method can almost always outperform baselines with a large margin, especially in long-term scenarios. It is also worth noting that final ADE and FDE values of our method SMART($c_{best}$) are at least 23\% and 39\% lower than that of others across all tables. 
We also observe that SMART($c_{best}$) provides better results than SMART($c_{random}$). This is due to the fact that SMART($c_{random}$) randomly samples trajectory behavior labels thus ignores the data distribution. In contrast, SMART($c_{best}$) is able to capture the diversity for particular label through CVAEs. Although other methods~\cite{social_gan,MATF,DEISRE} are also able to generate diverse trajectories, they have to sample a significant number of trajectories to get a predictions (or driver intents) exhibiting different behaviors. We later show in  Fig.~\ref{fig:ade_valid} that our method is able to model data distribution more effectively, e.g. achieves comparable/better results with less samples.
%We expect $c_{best}$ to give us the best performance as unlike other methods\cite{social_gan,MATF,DEISRE} where multi-modal diverse predictions are captured through a latent vector and are dependent on sampling a significant number of trajectories to get a predictions (or driver intents) exhibiting different behaviors, we split the diversity in the data through trajectory specific labels and capture diversity for a particular label through our CVAEs.~\buyu{This sounds very unfair actually..} 
% Hence, when computing negative log likelihood using $c_{random}$ sampling, the likelihood of our predictions with the ground truth is not the best in Tab.~\ref{tab:argoF} for the same reason mentioned earlier. But, computing the best of likelihood with respect to each label ($c_{best}$) yields the best results indicating the networks ability in capturing the data distribution. 

In Fig.~\ref{fig:ade_valid}(left), we show variation of ADE/FDE values with increasing number of samples in ArgoF. We observe that our method performs significantly better even with lower number of samples compared to baselines, which again supports our claim that methods like~\cite{social_gan} requires much more samples to even to get comparable performance with our method reported in Tab.~\ref{tab:argoF}(left). Fig.\ref{fig:ade_valid}(right) shows number of valid predictions produced across all datasets. Here, the validity is computed based on whether the predicted outputs lie within the road regions. Compared to baselines, our method is more likely to generate output predictions that satisfied context constraints. 

We also provide analysis on time complexity of existing methods in Tab.~\ref{tab:mean_runtime}. Without sacrificing the performance, our SMART always gives constant inference time with increasing number of agents. %Although efficient as MATF~\cite{MATF} seems to be, it actually provides poor experimental results and such efficicy is mainly due to the simple and optimized operations in MATF.

To demonstrate the effectiveness/informativeness of our {\bf simulated dataset}, we further conduct experiments and report numbers in Tab.~\ref{tab:argoT}. We test all methods on ArgoT test set. P-ArgoT in this table denotes that corresponding models are trained on P-ArgoT and fine-tuned on ArgoT training set. There are two main observations. Firstly, our methods that initialized with simulated data clearly achieve much better performance. Such significant performance boost indicates the benefits of augmenting a dataset with diverse trajectories. Secondly, such boost is missing in other methods. We argue that this might be attributed to the ability of the other methods in capturing the diversity in a wrong fashion (See Figure 3 in supplementary). For instance, S-GAN\cite{social_gan} is unaware of the scene context, hence when initialized with a model trained on diverse trajectories, the outputs are more spread thus leads to lower performance with fixed number of samples. Although MATF\cite{MATF} includes scene context in its predictions, it has poor capability in producing multimodal outputs where most of the predictions are biased towards behavior of particular type (See Figure 3. \cite{MATF}). To provide further analysis, and to show that training on simulated data improves diversity for other methods we report numbers evaluated on non-straight trajectories in Tab \ref{tab:argoT_no_straight}. As observed, other methods perform significantly better when when initialized with simulated model. This observation further demonstrates that our method is able to capture the diversity strongly coupled with the scene.

% We also conduct experiments to showcase the generalization of our simulated dataset. Specifically, we train models on either on simulated or real ArgoT dataset and test them on KITTI. Our results in Tab.~\ref{tab:kitti} show that models trained with our simulated but diverse dataset actually generate better results than that trained with real dataset. Such observation further shows that our simulated dataset generalizes better in general and has not been overfitted to particular dataset.

% we show predictions numbers of our method tested in KITTI\cite{KITTI} by training our method on their respective datasets. 

\begin{figure}[t]
    \centering
    % \begin{subfigure}{0.24\textwidth}
    %     \includegraphics[width=\linewidth, height=\linewidth]{./images/results_fig/5.png}
    %     \caption{}
    % \end{subfigure}
    \begin{subfigure}{0.24\textwidth}
        \includegraphics[width=\linewidth, height=\linewidth]{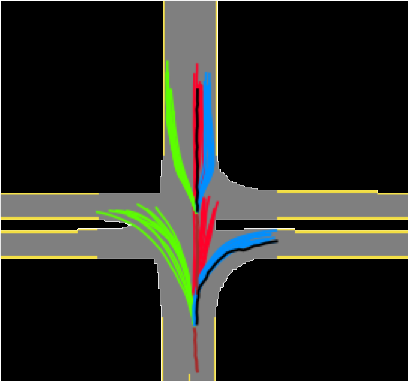}
        \caption{}
    \end{subfigure}
    % \begin{subfigure}{0.24\textwidth}
    %     \includegraphics[width=\linewidth, height=\linewidth]{./images/results_fig/9_multi.png}
    %     \caption{}
    % \end{subfigure}
    % \begin{subfigure}{0.24\textwidth}
    %     \includegraphics[width=\linewidth, height=\linewidth]{./images/results_fig/11_fail.png}
    %     \caption{}
    % \end{subfigure}
    \begin{subfigure}{0.24\textwidth}
        \includegraphics[width=\linewidth, height=\linewidth]{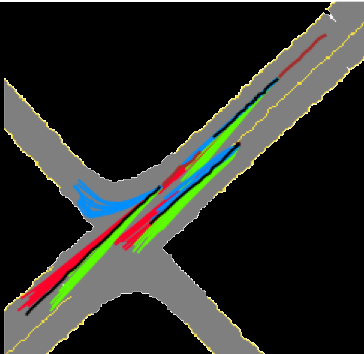}
        \caption{}
    \end{subfigure}
    \begin{subfigure}{0.24\textwidth}
        \includegraphics[width=\linewidth, height=\linewidth]{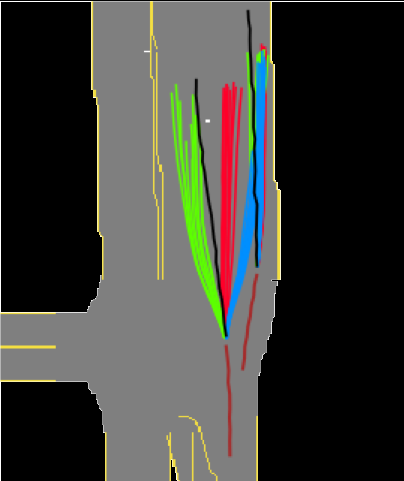}
        \caption{}
    \end{subfigure}
    % \begin{subfigure}{0.24\textwidth}
    %     \includegraphics[width=\linewidth, height=\linewidth]{./images/results_fig/argoF_1.png}
    %     \caption{}
    % \end{subfigure}
    \begin{subfigure}{0.24\textwidth}
        \includegraphics[width=\linewidth, height=\linewidth]{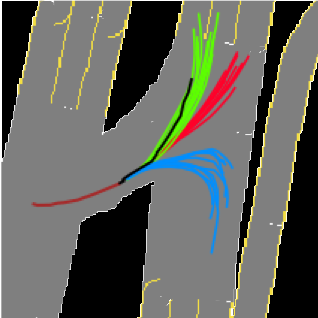}
        \caption{}
    \end{subfigure}
    
    \caption{Example predictions of SMART. The past trajectory and GT are visualized in brown and black lines. Red, blue and green lines are predictions sampled with different trajectory labels $c_i$ given as input. From left, multi-agent prediction outputs from simulated dataset P-ArgoT, ArgoT, ArgoT and ArgoF datasets. (a),(b) and (c) show simultaneous multi-agent multimodal outputs. (d) shows a failure case where some of the predicted trajectories are aligned in opposite to the direction of road. However, we argue that such traffic rules might be hard to obtain with only top view map information.}
%     \caption{Example predictions of SMART. The past trajectory and GT are visualized in brown and black lines. Red, blue and green lines are predictions sampled with different trajectory labels $c_i$ given as input. {\bf Top Row:} Predictions on simulated dataset (P-ArgoT). {\bf Bottom row:} Predictions on ArgoT and ArgoF. (a): lane change after turning with single agent.  (b),(c),(e) and (f) show simultaneous multi-agent trajectory prediction outputs.
% %(b),(c),(e) and (f) show simultaneous multi-agent trajectory prediction outputs. (b) and (g) show output predictions where the network correctly predicted right turn as one of the possibilities as in ground truth. (f) shows an interesting output when one agent performs a lane change and our predicted trajectories (shown in green) correctly predict the maneuver. In (b) while the network correctly predicts diverse paths for one agent, for the other vehicle which almost crossed the intersection it only predicts possible maneuvers in the straight direction showing strong conditioning of the scene context with the predictions. Same applies to output predictions in (a) where the vehicle started the turn. 
% (d) and (h) are failure cases where some of the predicted trajectories are aligned in opposite to the direction of road. However, we argue that such traffic rules might be hard to obtain with only top view map information.
%     }
    \label{fig:results_fig}
\end{figure}

\subsubsection{Qualitative Results}
We give some example predictions of our method in Fig.~\ref{fig:results_fig}. In general, our predictions align well with scene context and obey traffic rules in most situations. 

\subsubsection{Wasserstein diversity metric} To quantify the diversity of the simulated dataset, we introduce a novel diversity metric based on Wasserstein distances and showcase our results on both real and simulated data. Firstly, we normalize the trajectories such that it starts at the origin and ends at some x coordinate. We use a trajectory with zero acceleration $(\ddot x=0)$ and zero deviation from the x axis $(y=0)$ as a reference trajectory for comparison. We define two metrics $y$ (deviation from x axis) and $\ddot x$ (deviation from zero acceleration) Wasserstein. A higher Wasserstein metric indicates a higher deviation from the reference trajectory. Tab.~\ref{tab:argoF}(right) shows the Wasserstein metric between real and simulated data for two different datasets. Tracklets in KITTI~\cite{KITTI} and Argoverse~\cite{argoverse} generally move in straight directions with very minimal turns indicating a very low diversity. In contrast, our simulated trajectories are more diverse with agents executing turns whenever possible, going hand in hand with the higher diversity in Tab.~\ref{tab:argoF}(right).

\section{Conclusion}
In this paper, we have addressed  data diversity and model complexity issues in multiple-agent trajectory prediction. We first introduced a new simulated dataset that includes diverse yet realistic trajectories for multiple agents. Further, we propose SMART, a method that simultaneously generates trajectories for all agents with a single forward pass and provides multimodal, context-aware SOTA predictions. Our experiments on both real and simulated dataset show superiority of SMART over existing methods in terms of both accuracy and efficiency. In addition, we demonstrate that our simulated dataset is diverse and general, thus, is useful to train or test prediction models.

\bibliographystyle{splncs04}
\bibliography{references}
\clearpage
\setcounter{section}{0}
\setcounter{figure}{0}
\setcounter{table}{0}
% \author{}
% \title{Random}
% \maketitle

% \titlerunning{SMART Supplementary Material}
% \authorrunning{S. N N, et al.}
\pagestyle{headings}
% \phantom{.}  %necessary to add space on top before the title
\vspace{1cm}
\begin{center}
{\LARGE \bf Supplementary Material}\\[1cm]    
\end{center}

% \vspace{1cm}
\begin{abstract}
In this supplementary, we provide more details for our proposed simulator in Sec.~\ref{sec:sim} and for the proposed method for trajectory prediction, SMART in Sec.~\ref{sec:smart}. We further provide quantitative analysis and qualitative comparisons with a few prior works in Sec.~\ref{sec:results}. The accompanying video includes an overview of the method and qualitative visualizations of our predictions.
\end{abstract}
% \section{TODO}
% \begin{itemize}
%     \item Comparison figures on perspective view images for baselines and ablative baselines
%     \item Show improvement in output diversity with the simulated data using S-GAN (Video or paper).
%     \item Add captaions
%     \item Tighter crop top view images
% \end{itemize}

\section{Further Details for Simulator}~\label{sec:sim}
In this section, we provide more details about reference velocity sampling, low-level controller and also specify the range of values for IDM \cite{Treiber2000CongestedTS} parameters used in our simulation.
% ~\buyu{Give a simple flowchart here so that we can remind the reviewers the pipeline of our simulator.}
% ~\buyu{A short summary of what has been included in supple. is needed. Please also check the paper to make sure all we have referred are included in this file.}

\subsection{Velocity Sampling:}
\begin{figure}[!!t]
    % \centerArgoing
    \begin{tabular}{cc}
    \includegraphics[width=0.6\linewidth]{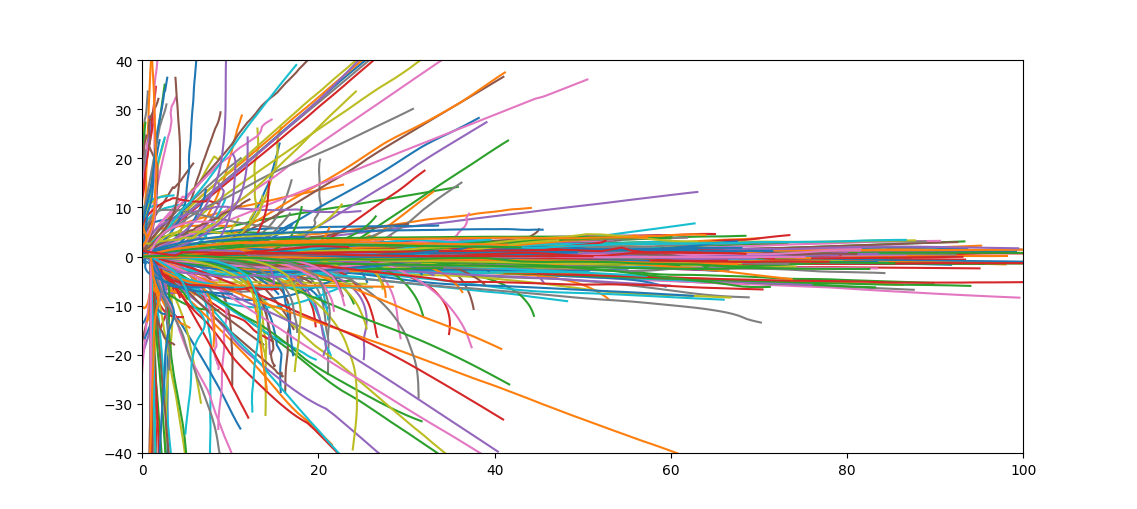} &
    \includegraphics[width=0.35\linewidth]{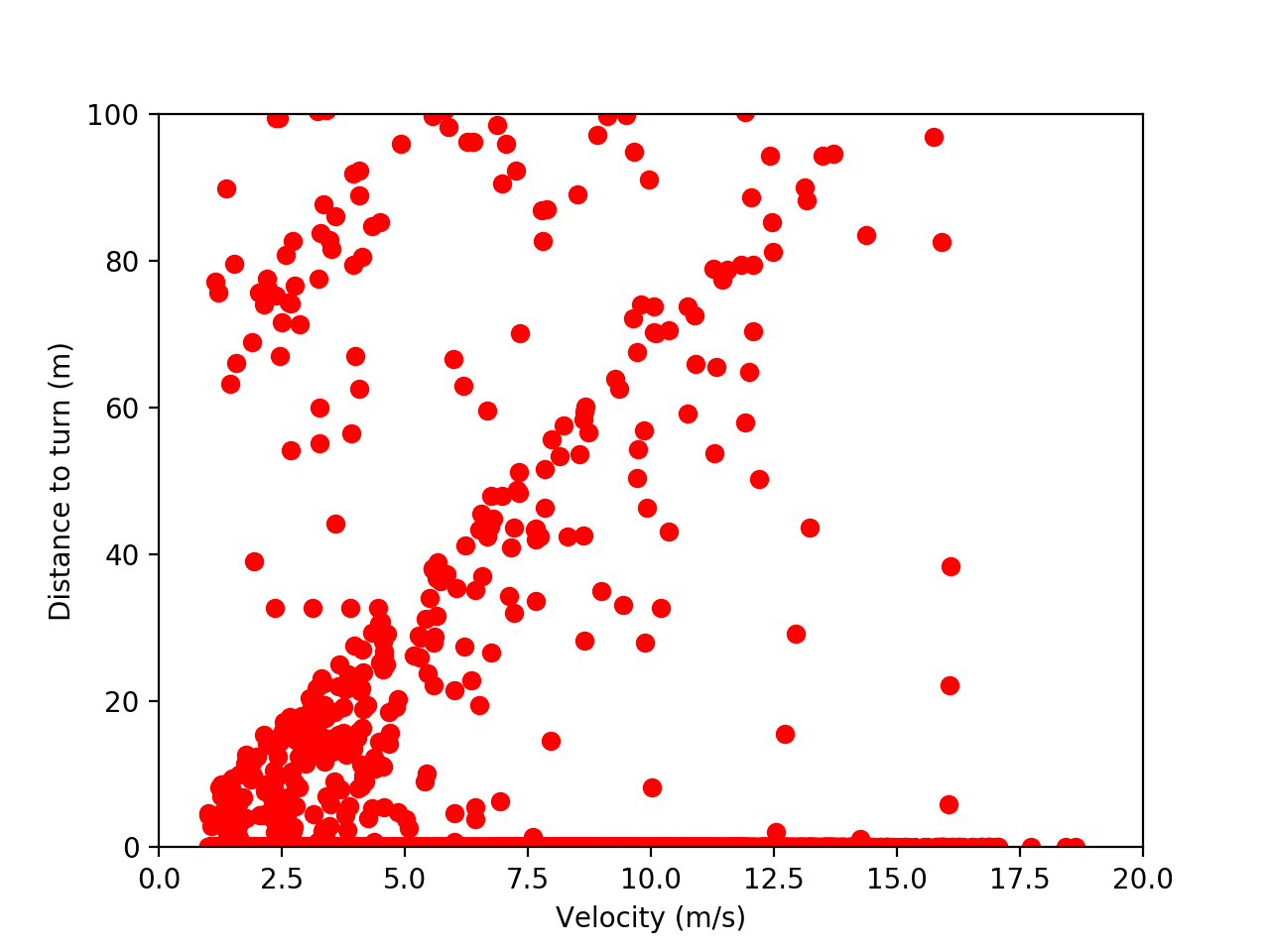}\\
    \end{tabular}
    \caption{{\bf Left:} Shows several \emph{\{left,right and straight\}} trajectories executed by vehicles in real world on Argoverse tracking (ArgoT) data sampled at 1Hz. {\bf Right:} Velocity \emph{vs} Distance to turn plot for tracklets in Argoverse tracking for the entire dataset. For plotting purposes we plot distance to turn to be zero for straight maneuvers. Here each dot represents a velocity profile for 7 seconds.}
    \label{fig:left_right_dist_vel}
\end{figure}

Figure \ref{fig:left_right_dist_vel} shows several real trajectories ($x$ and $y$ axes represent distance travelled in meters in longitudinal and lateral directions) and a plot between distance before turn and average velocity for real trajectory samples. We calculate distance before turn as the distance travelled by the vehicle before it starts executing a turn maneuver and average velocity is the mean velocity through the course of the trajectory before turn.
% ~\buyu{How do you compare the distance and velocity?}. 
Interestingly, we found distance before taking turns %as a reliable feature for sampling velocity profiles for turning maneuvers.
is highly correlated to average velocity. Specifically,
we can see a trend of decrease(or slowing down) in average velocity
%(or slowing down before an intersection) among vehicles as we move close towards making a turn maneuver. 
while approaching an intersection with an intention of making a turn maneuver.
To this end, we identify distance to intersections as useful feature  
%Hence, identifying closest profiles based on such features helps 
and demonstrate that it helps in mimicking the real data. We label every velocity profile from the real data with a value of distance before turn or average velocity, for turn and straight maneuvers respectively. Here, by velocity profile we mean a series of vehicle velocities at every timestep for the simulation period. We identify the nearest neighbor velocity profile in the real data using the above mentioned features with the values from the simulated vehicle. We use the identified velocity profile as reference for the simulated vehicle to achieve at every timestep. We also add gaussian noise with zero mean and unit variance to add diversity in the sampled velocity profiles.
% More specifically, we identify the closest velocity profile based on such feature and each vehicle is initialized with a velocity from the sampled velocity profile. We also add gaussian noise with zero mean and unit variance to add diversity in reference velocities.
% ~\buyu{How many samples are we working on here for the right handside figure? Is this the data distribution on ArgoT or ArgoP?}

\subsection{Low Level Controller:}
Low-level controller simulates the desired behavior governed by vehicle dynamics module. It takes input from maneuver identification, IDM\cite{Treiber2000CongestedTS} and MOBIL\cite{mobil}, and produces state changes for the simulated vehicle. It consists of longitudinal and lateral proportional controllers that give out required velocity commands. The lane centerline is used as the reference trajectory for the simulated vehicle to follow.
The velocity obtained from the lateral controller is converted to appropriate steering commands that helps in tracking the reference trajectory. Let $v$ be the current velocity of the vehicle, $x_{lateral}$ be the lateral position from the lane and $v_{lateral}$ be the lateral velocity then steering angle $\phi$ is obtained through the following set of equations:
\begin{equation}
    v_{lateral} = - kp_{lateral} * ( x_{lateral} + \epsilon )
\end{equation}
\begin{equation}
    \psi_{req} = \arcsin(\frac{v_{lateral}}{v})
\end{equation}
\begin{equation}
    \psi_{ref} = \psi_{future} + \psi_{req}
\end{equation}
\begin{equation}
    \Dot{\psi} = kp_{heading} * \psi_{ref}
\end{equation}
\begin{equation}
    \phi = \arctan(\frac{L}{v}\Dot{\psi})
    \label{eq:steering_cnvt}
\end{equation}
where $kp_{lateral}$ and $kp_{heading}$ are controller parameters, $L$ represents length of the vehicle and $\epsilon$ acts as an offset noise in tracking the lane. $\psi_{req}$ is the heading that needs to be compensated for aligning with the lane center, while $\psi_{future}$ is the required heading that needs to be achieved for future timesteps. A heading controller provides a heading rate $\Dot{\psi}$ for the given reference heading $\psi_{ref}$. Equation \ref{eq:steering_cnvt} calculates the steering angle based on current velocity $v$, vehicle length $L$ and heading rate $\Dot{\psi}$.

\subsection{Intelligent Driver Model:}
Tab \ref{tab:idm_params} shows values and sample space for parameters in Intelligent Driver Model\cite{Treiber2000CongestedTS}. We sample these parameters randomly to increase diversity of driving patterns.
%during simulation 
% to generate simulations with different levels of aggressiveness~\buyu{What does different level of aggressiveness mean? More aggressive mean the driver is more aggressive or what? Or you actually mean "to increase diversity of driving patterns"?}.

\begin{table}[!!t]
    \centering
    \caption{Parameters for Intelligent Driver Model.}
    \resizebox{0.6\linewidth}{!}{\begin{tabular}{|c|c|c|}
    \hline
         Parameter & Values & Units\\
    \hline         
         Desired velocity $v_o$ & Reference profile & $m/s$\\
         Free acceleration exponent $\delta$ & $4.0$ & -\\
         Safety time gap $T$ & $range(0.5, 2.5)$ & $s$\\
         Desired safety distance $s_o$ & $range(0.5, 4.0)$ & $m$\\
         Comfortable acceleration $a$ & $1.5 \pm 0.5$ & $m/s^2$\\
         Comfortable deceleration $b$ & $2.0 \pm 0.5$ & $m/s^2$\\
    \hline
    \end{tabular}}
    \label{tab:idm_params}
\end{table}

\vspace{1cm}
\section{Further Details for SMART}~\label{sec:smart}

\vspace{-0.4cm}\subsection{Network Architecture:}
In this section, we provide more details of our network architecture. 

\vspace{-0.4cm}\subsubsection{Latent Encoder:} It takes concatenated past and future trajectories and a corresponding trajectory label as input. The embedding layer and LSTM contain 16 dimensions. The fully connected layer has 512 and 128 units that produce 16 dimensional $\mu$ and $\sigma$ as outputs.

% The input trajectory is passed through an embedding layer to output a 16 dimensional vector at every timestep. The embedded vector passes through a LSTM layer of 16 dimensions. The output hidden vectors from the LSTM are concatenated along with one hot trajectory label. It is then fed through fully connected layers of 512,128 dimensions to produce 16 dimensional $\mu$ and $\sigma$ as outputs.~\buyu{I feel like this paragraph alreay exists in main submission.}

\vspace{-0.4cm}\subsubsection{Convolutional Encoder:} Our encoder receives input of HxWx25 dimension. It consists of 6 convolutional layers with first layer being 3D convolution and followed by 2D ones. The number of filters are 16,16,32,32,64 and 64 from the very first to the sixth layer, respectively, with alternativing stride 1 and stride 2. We set the kernel size to 1x1x4 in the first layer and that of the remaining layers to 3x3. 
% alternative stride 1 and stride 2 convolutions. The Convolutional Encoder has a total of 6 convolutional layers with first layer being 3D convolution and followed by 2D. The following are the number of filters for each layer 16,16,32,32,64,64 respectively. The kernel size for 3D convolution is of 1x1x4 while for 2D convolutions we have a standard kernel size of 3x3.

\vspace{-0.4cm}\subsubsection{ConvLSTM Decoder} It consists of alternating ConvLSTMs and 2D transposed convolutional layers. The size of hidden state or output filters in every pair of ConvLSTMs and transposed convolutions are 64,32 and 16, respectively. Convolutions share the same kernel size 3x3. And the transposed convolutions have a stride of 2. We add skip connections from encoder layers 2 and 4 to corresponding second and third ConvLSTM layers in the decoder.

Finally, a ConvLSTMs with state-pooling operation is further put to the end of decoder. It has a hidden state of 16 channels with 1x1 kernel size. We also concatenate features from first 3D convolution before feeding it to ConvLSTM layer. In addition, we include one last convolution layer that generates 2 channels with 1x1 kernel size as the output layer. 
% All weights are initialized based on variance scaling with a scale value of 0.1~\buyu{If you prefer to keep the last sentence, maybe you should state it in the beginning of this section.}.

\subsection{Learning Details}
The models are trained using Adam optimizer with a learning rate of 0.008 and a batch size of 6. The model is trained on ArgoF for 10 epochs and 400 epochs on both ArgoT and P-ArgoT. We train the models at the trajectory frequency of 5hz and interpolate the results at the desired frequency. In order to avoid exploding gradients, we apply gradient clipping with L2 norm of 1.0. Further, during the training procedure, we augment the data by randomly rotating the scene and trajectories to reduce over-fitting. All models are implemented using Tensorflow 2.0 and trained with a NVIDIA RTX 2080Ti GPU.

\subsection{Future work}
SMART methods capabilities can be extended by incorporating traffic rules to reduce the number of invalid trajectories that span in the wrong direction (Figure 6d main paper). Furthermore, explicitly modelling interactions among multiple agents improve predictions and reduce invalid trajectory collisions with other agents in the scene. 
\section{Further Results}\label{sec:results}
We provide more qualitative and quantitative analysis in this section.

% \sriram{Need to combine the figures and place them correctly}
\vspace{-0.4cm}\subsubsection{Realism of Simulated Data} 
\begin{figure}[h]
    \centering
    \includegraphics[width=0.6\linewidth]{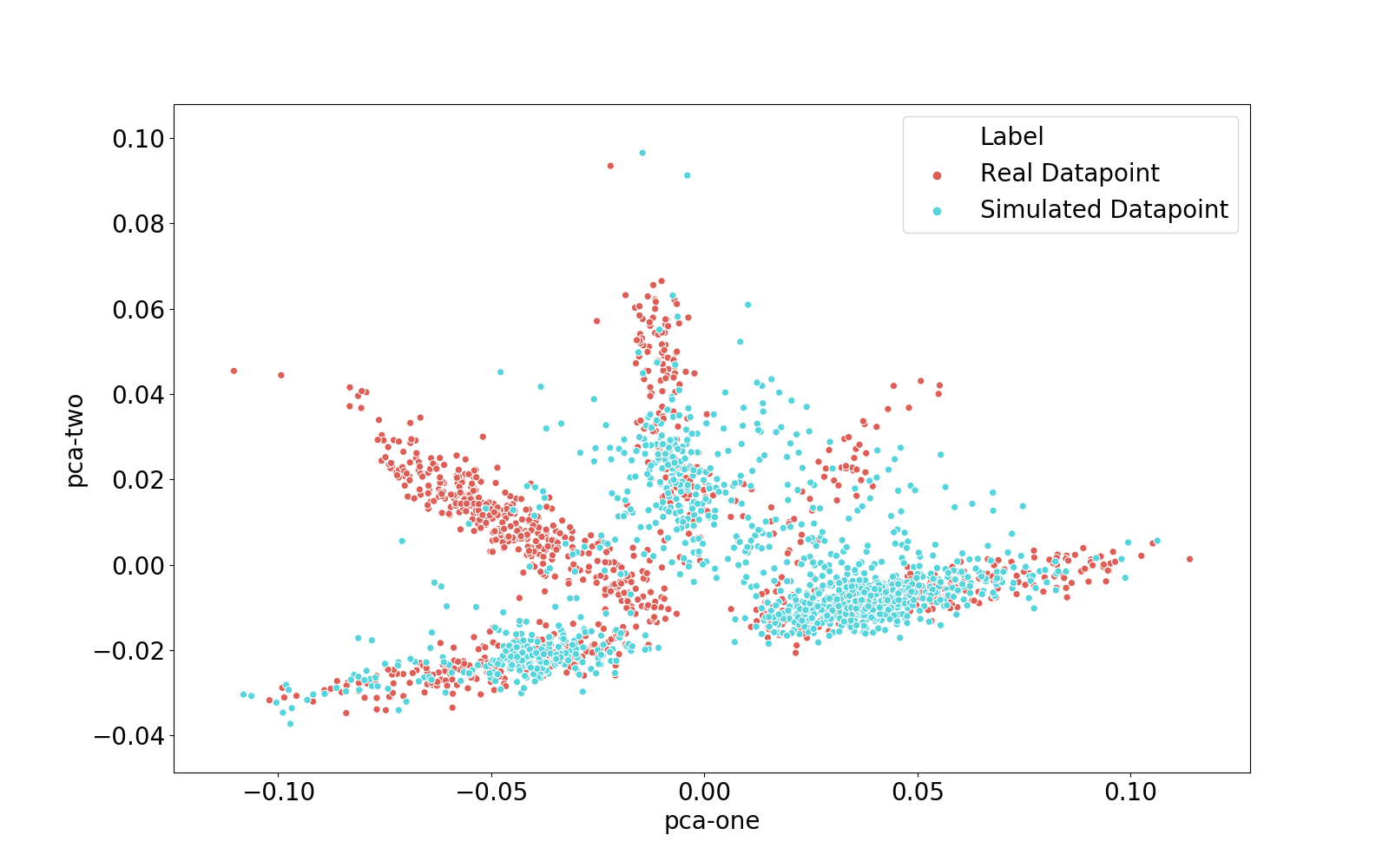}
    \caption{Qualitative plot of PCA performed on 1000 random real and simulated trajectories from Argo Tracking}
    \label{fig:pca_argoT}
\end{figure}

To evaluate the realism of our simulated dataset, we perform PCA on a random set of real and simulated trajectories, followed by a Gaussian KDE on the PCA-transformed real trajectories. The calculated log likelihood for both real and simulated data on real fitted KDE for 1000 random datapoints are 2.25 and 2.19. This indicates that the simulated distribution falls very close to the real one. A qualitative plot of real and simulated trajectories after PCA transformation is shown in Figure \ref{fig:pca_argoT}.

\vspace{-0.4cm}\subsubsection{Visualization for diversity} 
\begin{figure}[!!t]
    \centering
    \begin{tabular}{c|c|c|c}
        
        Train set& MATF-GAN\cite{MATF} & S-GAN\cite{social_gan} & SMART\\
        \hline
        ArgoT &
         \raisebox{-.5\height}{\includegraphics[width=0.15\linewidth]{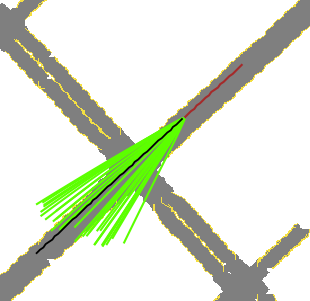}}&
         \raisebox{-.5\height}{\includegraphics[width=0.15\linewidth]{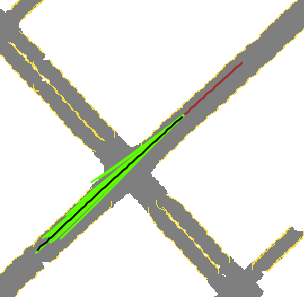}}&
         \raisebox{-.5\height}{\includegraphics[width=0.15\linewidth]{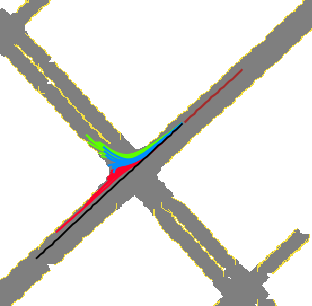}}
         \\
         \hline
         P-ArgoT&
         \raisebox{-.5\height}{\includegraphics[width=0.15\linewidth]{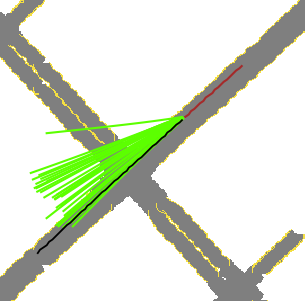}}&
         \raisebox{-.5\height}{\includegraphics[width=0.15\linewidth]{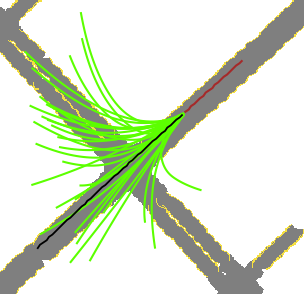}}&
         \raisebox{-.5\height}{\includegraphics[width=0.15\linewidth]{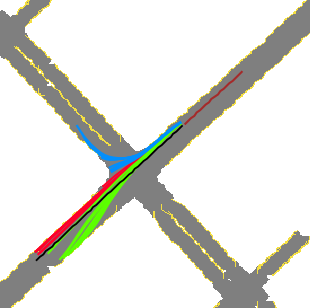}}
         \\
         \hline
    \end{tabular}
    
    \caption{Qualitative comparison of methods tested on Argo Tracking(ArgoT) with only training on ArgoT and training on P-ArgoT and fine tuning on ArgoT. Significant improvement in the diversity of prediction results can be observed in S-GAN\cite{social_gan} when the model is initialized using our simulated data.}
    \label{fig:fine_tune_results}
\end{figure}

Figure \ref{fig:fine_tune_results} shows comparison of prediction outputs for methods trained only on real data and methods trained on simulated data and fine-tuned on real data. Clear improvement in the diversity can be observed with S-GAN\cite{social_gan}. But it does not capture such diversity with the scene context attributing to higher deviation with the ground truth for the same number of samples while our method captures diversity coupled with the semantics in the right way enabling it to provide outputs towards different modes. 

\vspace{-0.4cm}\subsubsection{Results on KITTI} We also conduct experiments to showcase the generalization of our simulated dataset. Specifically, we train models either on simulated or real ArgoT dataset and test them on KITTI. Our results in Tab.~\ref{tab:kitti} show that models trained with our simulated but diverse dataset actually generate better results than that trained with real dataset. Such observation further shows that our simulated dataset generalizes better in general and has not been overfitted to particular dataset.

\begin{table}[h]
\caption{Quantitative results on KITTI dataset. `[ ]' denotes the training set.}
\centering
\resizebox{0.98\textwidth}{!}{
\begin{tabular}{c ||c|c|c|| c|c|c|| c|c|c|| c|c|c ||c|c|c}
\hline
{Model} & \multicolumn{3}{c||}{1.0(sec)} & \multicolumn{3}{c||}{2.0(sec)} & \multicolumn{3}{c||}{3.0(sec)} & \multicolumn{3}{c||}{4.0(sec)} & \multicolumn{3}{c}{5.0(sec)}\\ 
    \hline
    \multicolumn{16}{c}{KITTI Dataset $\|$ ADE $\vert$ FDE $\vert$ NLL $\|$ }\\
    \hline
    SMART $[$ArgoT$]$ $(c_{random})$ & 1.30 & 2.06 & 8.45 & 2.47 & 4.76 & 8.50 & 3.88 & 7.97 & 8.75 & 5.19 & 9.11 & 8.93 & 6.14 & 11.2 & 8.98\\
    SMART $[$PArgoT$]$ $(c_{random})$ & 1.14 & 1.77 & 5.33 & 2.12 & 3.94 & 6.03 & 3.29 & 6.55 & 6.61 & 4.53 & 8.69 & 7.13 & 5.88 & 9.92 & 7.58\\
    \hline
    SMART $[$ArgoT$]$ $(c_{best})$ & 1.22 & 1.92 & 7.10 & 2.30 & 4.38 & 7.42 & 3.58 & 7.28 & 7.87 & 4.75 & 8.10 & 8.174 & 5.65 & 9.88 & 8.28\\
    SMART $[$PArgoT$]$ $(c_{best})$ & {\bf 1.03} & {\bf 1.56} & {\bf 5.29} & {\bf 1.89} & {\bf 3.50} & {\bf 5.95} & {\bf 2.94} & {\bf 5.77} & {\bf 6.56} & {\bf 4.06} & {\bf 7.60} & {\bf 7.08} & {\bf 5.32} & {\bf 8.71} & {\bf 7.49}\\
    \hline

\end{tabular}
}
\label{tab:kitti}
\end{table}

\vspace{-0.4cm}\subsubsection{Further Qualitative Results} 
\begin{figure}
    \centering
    % \begin{tabular}{ccccc}

    %      MATF-GAN\cite{MATF}&
    %      \includegraphics[width=0.20\linewidth]{./supp_results/matf_argoT/left/72.png}&
    %      \includegraphics[width=0.20\linewidth]{./supp_results/matf_argoT/straight/72.png}&
    %      \includegraphics[width=0.20\linewidth]{./supp_results/matf_argoT/right/72.png}&
    %      \includegraphics[width=0.15\linewidth, height=0.13\linewidth]{./supp_results/matf_argoT/bev/72.png}
    %      \\
    %      S-GAN\cite{social_gan}&
    %      \includegraphics[width=0.20\linewidth]{./supp_results/sgan_argoT/left/72.png}&
    %      \includegraphics[width=0.20\linewidth]{./supp_results/sgan_argoT/straight/72.png}&
    %      \includegraphics[width=0.20\linewidth]{./supp_results/sgan_argoT/right/72.png}&
    %      \includegraphics[width=0.15\linewidth, height=0.13\linewidth]{./supp_results/sgan_argoT/bev/72.png}
    %      \\
    %      SMART &
    %      \includegraphics[width=0.20\linewidth]{./supp_results/ours_simArgoT/left/72.png}&  
    %      \includegraphics[width=0.20\linewidth]{./supp_results/ours_simArgoT/straight/72.png}&
    %      \includegraphics[width=0.20\linewidth]{./supp_results/ours_simArgoT/right/72.png}&
    %      \includegraphics[width=0.15\linewidth,height=0.13\linewidth]{./supp_results/ours_simArgoT/bev/72.png}\\
          
    % \end{tabular}

    \begin{tabular}{p{2cm}cccc}
        
         MATF-GAN\cite{MATF}&
         \raisebox{-.5\height}{\includegraphics[width=0.20\linewidth]{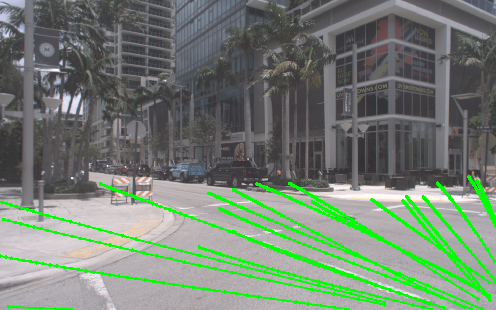}}&
         \raisebox{-.5\height}{\includegraphics[width=0.20\linewidth]{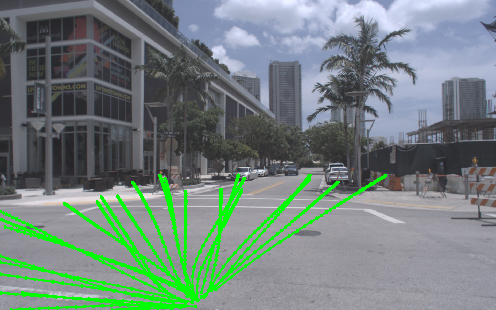}}&
         \raisebox{-.5\height}{\includegraphics[width=0.20\linewidth]{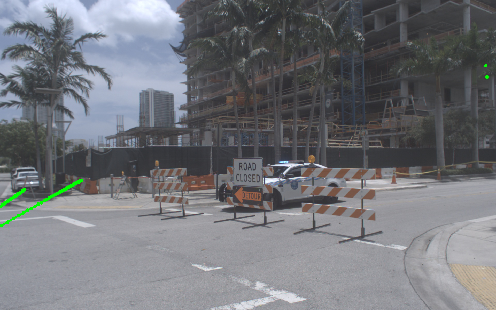}}&
         \raisebox{-.5\height}{\includegraphics[width=0.20\linewidth, height=0.13\linewidth]{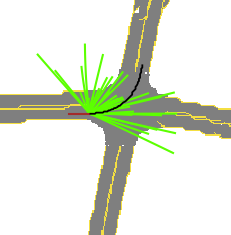}}
         \\
         S-GAN\cite{social_gan}&
         \raisebox{-.5\height}{\includegraphics[width=0.20\linewidth]{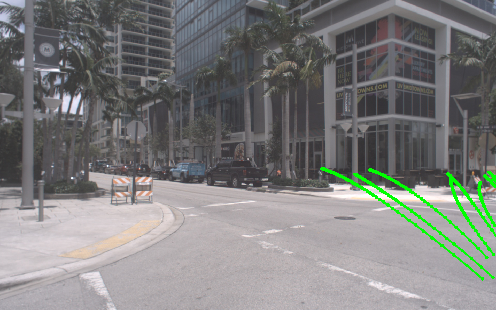}}&
         \raisebox{-.5\height}{\includegraphics[width=0.20\linewidth]{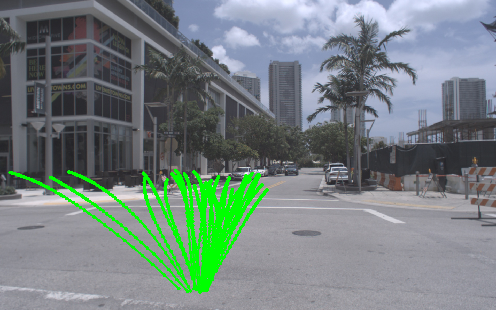}}&
         \raisebox{-.5\height}{\includegraphics[width=0.20\linewidth]{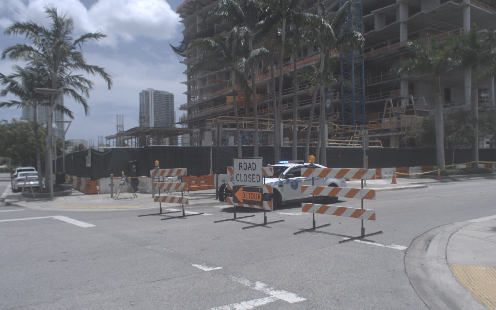}}&
         \raisebox{-.5\height}{\includegraphics[width=0.20\linewidth, height=0.13\linewidth]{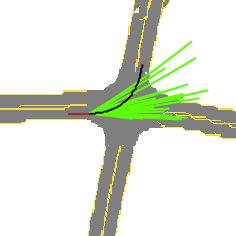}}
         \\
         SMART &
         \raisebox{-.5\height}{\includegraphics[width=0.20\linewidth]{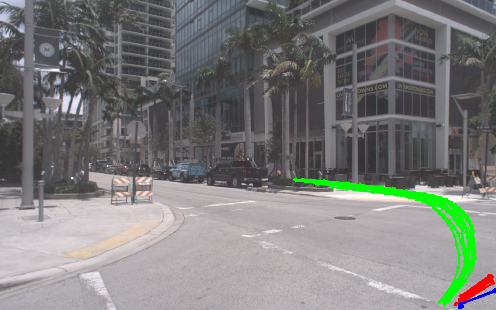}}&  
         \raisebox{-.5\height}{\includegraphics[width=0.20\linewidth]{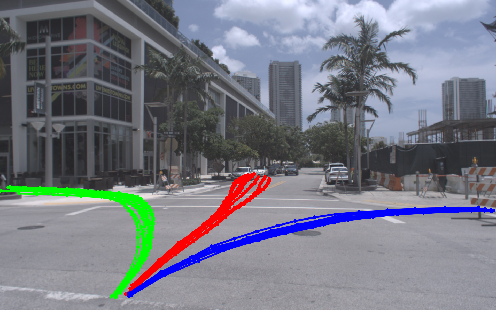}}&
         \raisebox{-.5\height}{\includegraphics[width=0.20\linewidth]{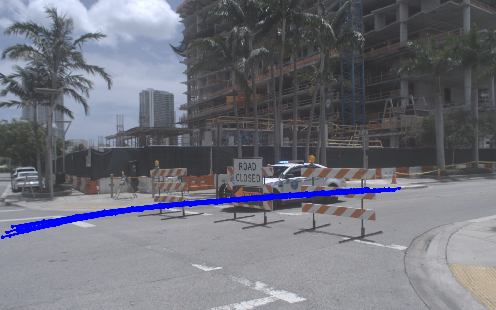}}&
         \raisebox{-.5\height}{\includegraphics[width=0.20\linewidth,height=0.13\linewidth]{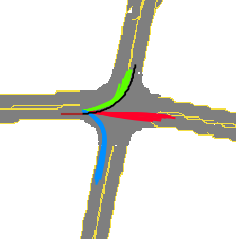}}\\
          \\
    \end{tabular}

    \begin{tabular}{p{2cm}cccc}

         MATF-GAN\cite{MATF}&
         \raisebox{-.5\height}{\includegraphics[width=0.20\linewidth]{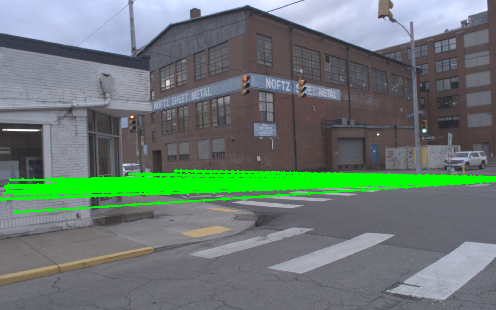}}&
         \raisebox{-.5\height}{\includegraphics[width=0.20\linewidth]{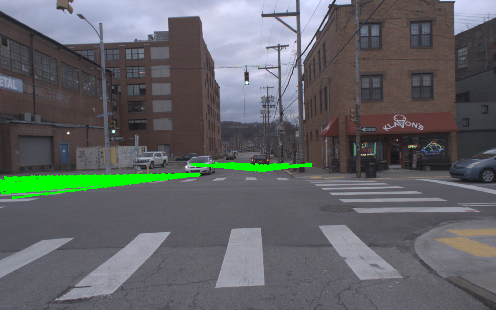}}&
         \raisebox{-.5\height}{\includegraphics[width=0.20\linewidth]{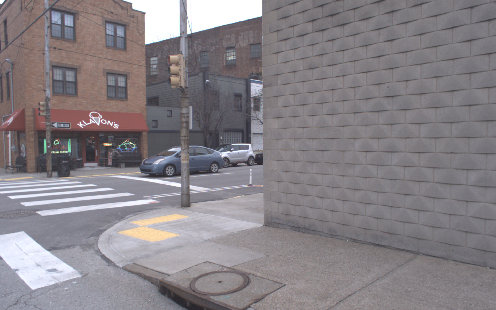}}&
         \raisebox{-.5\height}{\includegraphics[width=0.20\linewidth, height=0.13\linewidth]{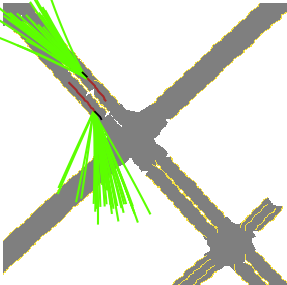}}
         \\
         S-GAN\cite{social_gan}&
         \raisebox{-.5\height}{\includegraphics[width=0.20\linewidth]{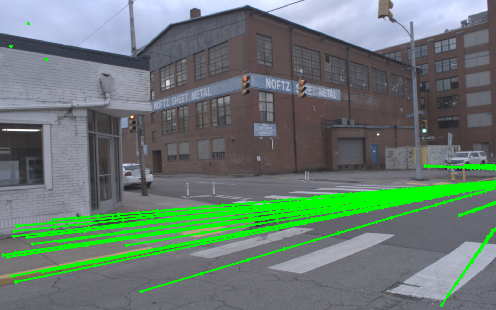}}&
         \raisebox{-.5\height}{\includegraphics[width=0.20\linewidth]{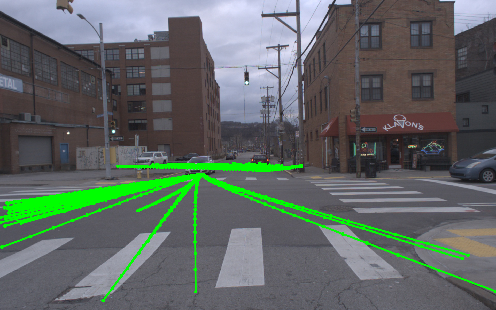}}&
         \raisebox{-.5\height}{\includegraphics[width=0.20\linewidth]{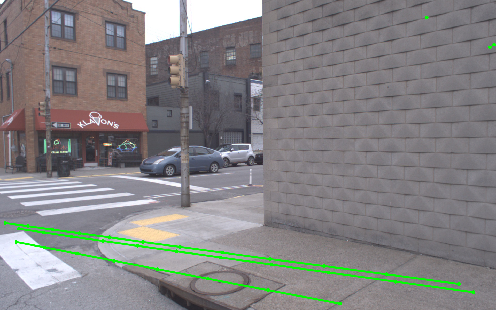}}&
         \raisebox{-.5\height}{\includegraphics[width=0.20\linewidth, height=0.13\linewidth]{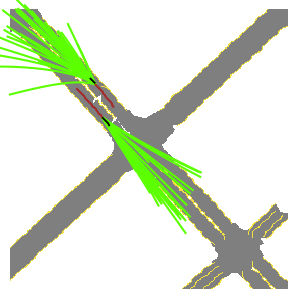}}
         \\
         SMART &
         \raisebox{-.5\height}{\includegraphics[width=0.20\linewidth]{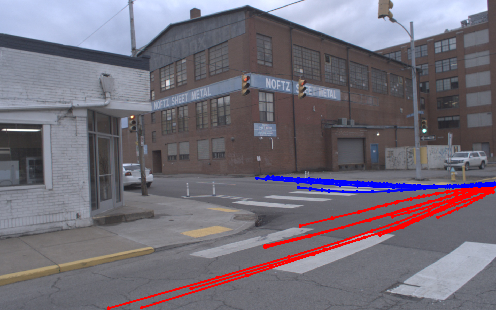}}&  
         \raisebox{-.5\height}{\includegraphics[width=0.20\linewidth]{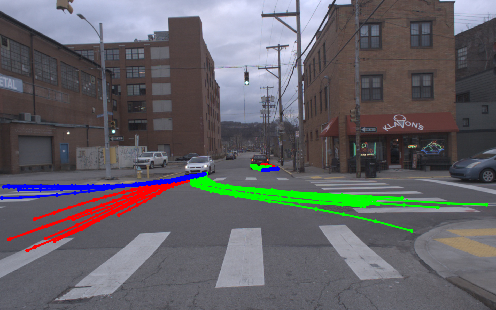}}&
         \raisebox{-.5\height}{\includegraphics[width=0.20\linewidth]{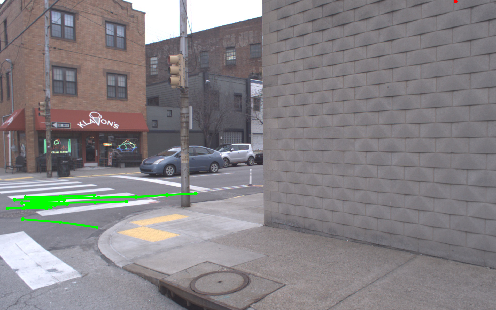}}&
         \raisebox{-.5\height}{\includegraphics[width=0.20\linewidth,height=0.13\linewidth]{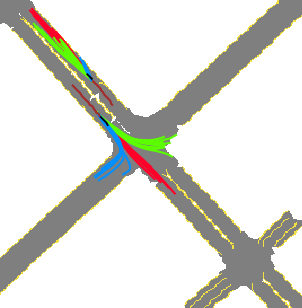}}\\
          \\
    \end{tabular}
    \begin{tabular}{p{2cm}cccc}
        
         MATF-GAN\cite{MATF}&
         \raisebox{-.5\height}{\includegraphics[width=0.20\linewidth]{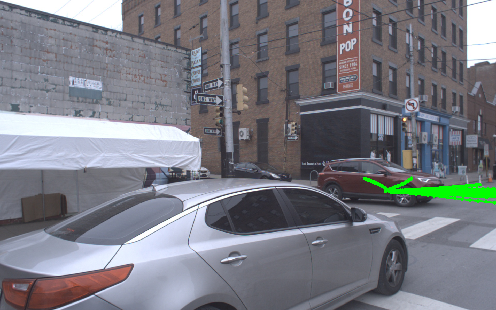}}&
         \raisebox{-.5\height}{\includegraphics[width=0.20\linewidth]{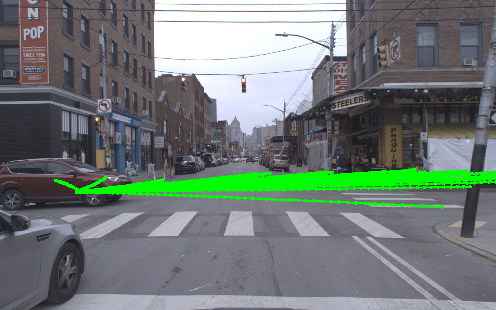}}&
         \raisebox{-.5\height}{\includegraphics[width=0.20\linewidth]{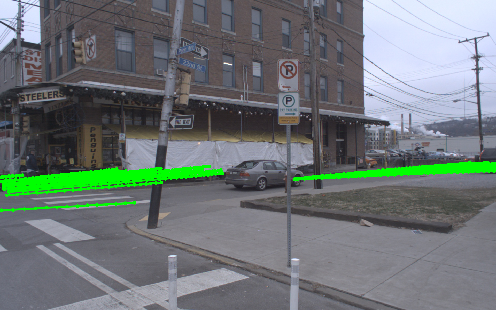}}&
         \raisebox{-.5\height}{\includegraphics[width=0.20\linewidth, height=0.13\linewidth]{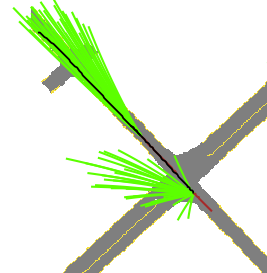}}
         \\
         S-GAN\cite{social_gan}&
         \raisebox{-.5\height}{\includegraphics[width=0.20\linewidth]{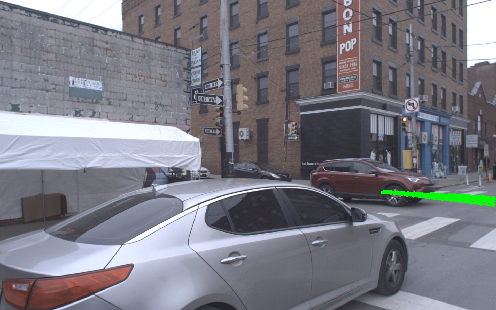}}&
         \raisebox{-.5\height}{\includegraphics[width=0.20\linewidth]{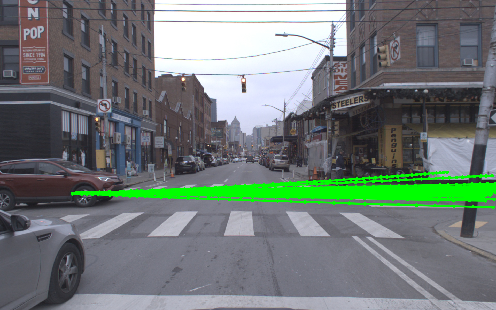}}&
         \raisebox{-.5\height}{\includegraphics[width=0.20\linewidth]{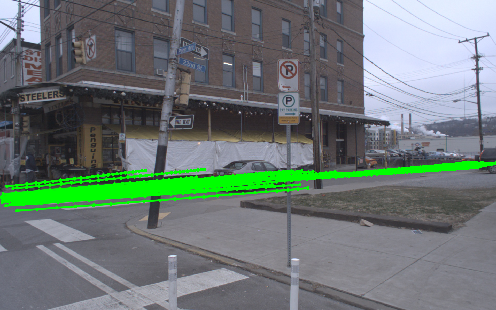}}&
         \raisebox{-.5\height}{\includegraphics[width=0.20\linewidth, height=0.13\linewidth]{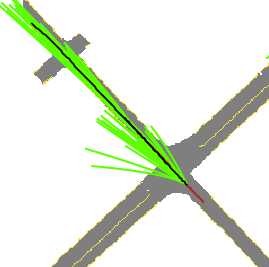}}
         \\
         SMART &
         \raisebox{-.5\height}{\includegraphics[width=0.20\linewidth]{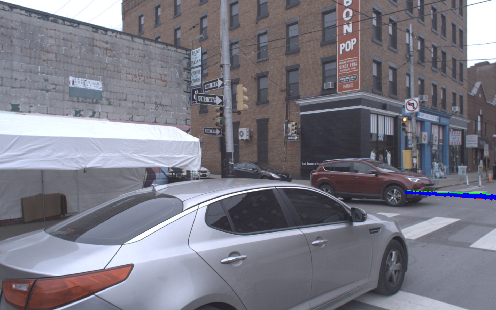}}&  
         \raisebox{-.5\height}{\includegraphics[width=0.20\linewidth]{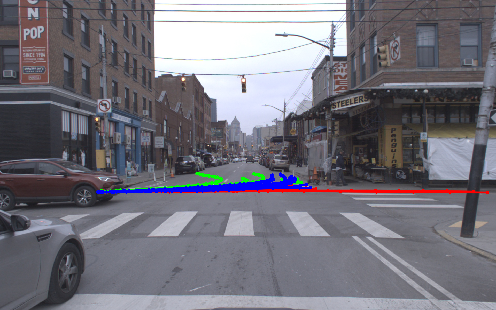}}&
         \raisebox{-.5\height}{\includegraphics[width=0.20\linewidth]{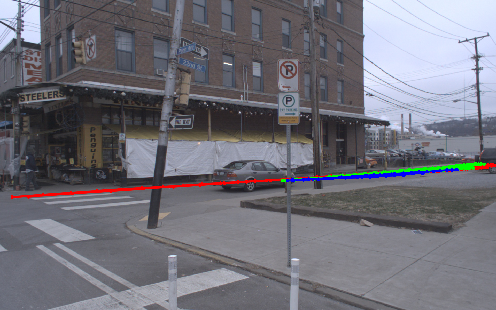}}&
         \raisebox{-.5\height}{\includegraphics[width=0.20\linewidth,height=0.13\linewidth]{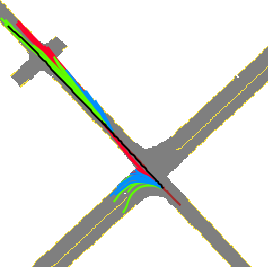}}\\
          
    \end{tabular}
    
    \caption{Qualitative comparison of SMART with other baseline methods. The figure depicts predicted outputs plotted on left, center, right camera images and top view images of the scene. Note that for SMART, red, blue and green trajectories show outputs with different trajectory labels that captures different modes in the output.}
    \label{fig:supp_results}
\end{figure}

\begin{figure}
    \centering
    \centering
    \begin{tabular}{p{2cm}cccc}

         MATF-GAN\cite{MATF}&
         \raisebox{-.5\height}{\includegraphics[width=0.20\linewidth]{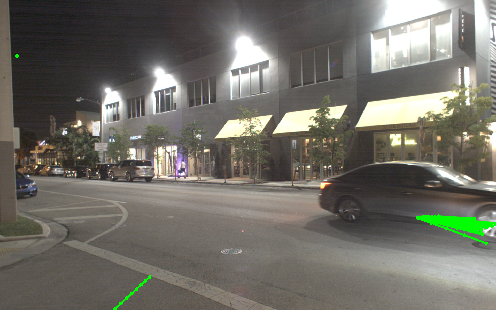}}&
         \raisebox{-.5\height}{\includegraphics[width=0.20\linewidth]{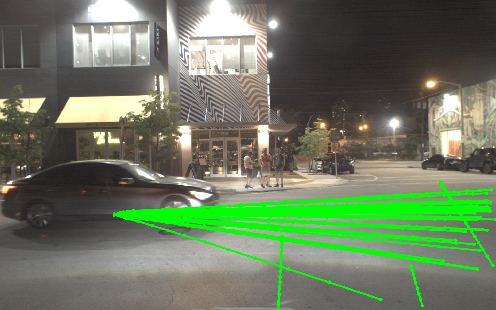}}&
         \raisebox{-.5\height}{\includegraphics[width=0.20\linewidth]{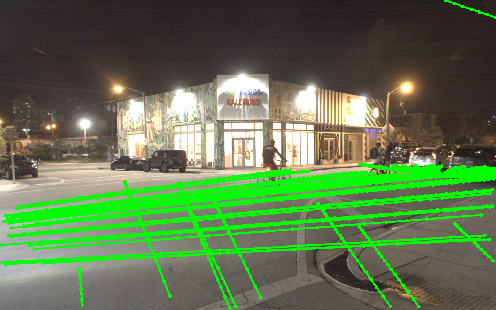}}&
         \raisebox{-.5\height}{\includegraphics[width=0.20\linewidth, height=0.13\linewidth]{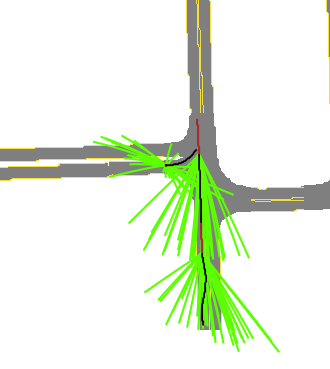}}
         \\
         S-GAN\cite{social_gan}&
         \raisebox{-.5\height}{\includegraphics[width=0.20\linewidth]{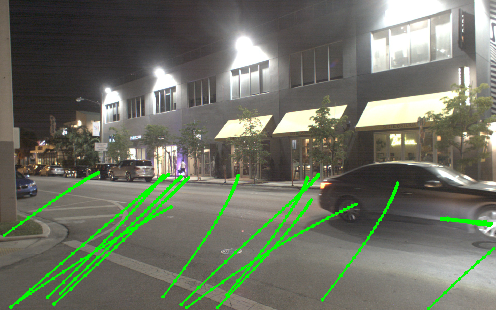}}&
         \raisebox{-.5\height}{\includegraphics[width=0.20\linewidth]{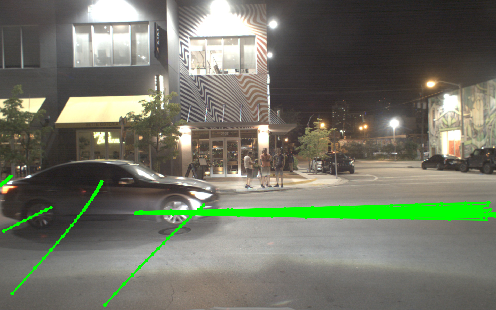}}&
         \raisebox{-.5\height}{\includegraphics[width=0.20\linewidth]{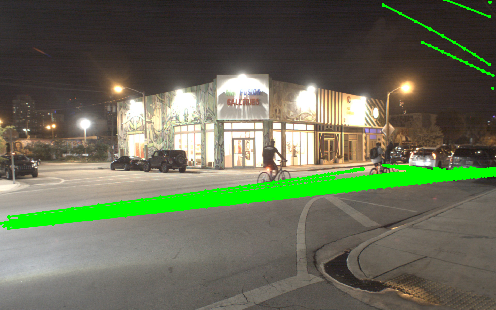}}&
         \raisebox{-.5\height}{\includegraphics[width=0.20\linewidth, height=0.13\linewidth]{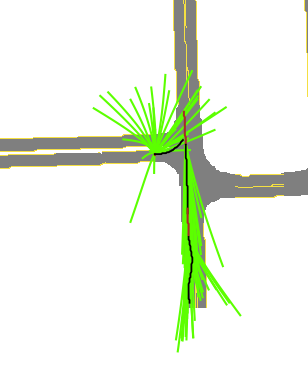}}
         \\
         SMART &
         \raisebox{-.5\height}{\includegraphics[width=0.20\linewidth]{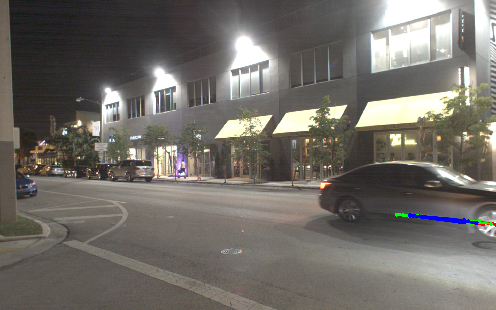}}&  
         \raisebox{-.5\height}{\includegraphics[width=0.20\linewidth]{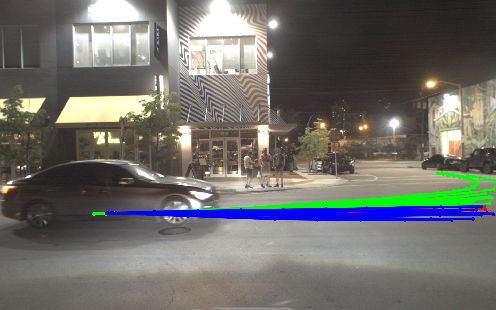}}&
         \raisebox{-.5\height}{\includegraphics[width=0.20\linewidth]{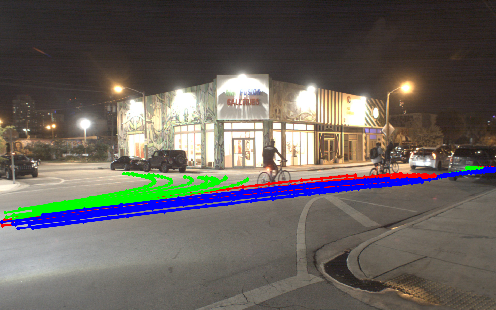}}&
         \raisebox{-.5\height}{\includegraphics[width=0.20\linewidth,height=0.13\linewidth]{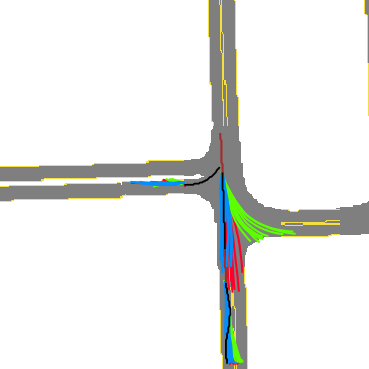}}\\
          
    \end{tabular}
    
    \vspace{0.2cm}
    
    \begin{tabular}{p{2cm}cccc}

         MATF-GAN\cite{MATF}&
         \raisebox{-.5\height}{\includegraphics[width=0.20\linewidth]{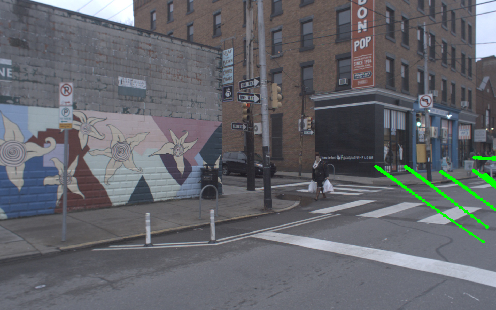}}&
         \raisebox{-.5\height}{\includegraphics[width=0.20\linewidth]{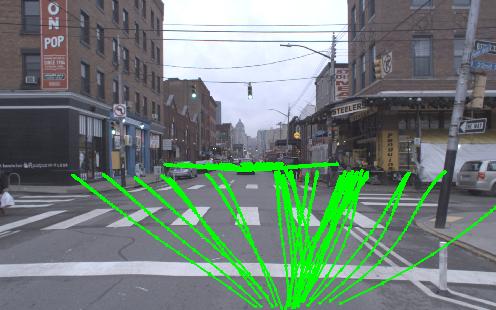}}&
         \raisebox{-.5\height}{\includegraphics[width=0.20\linewidth]{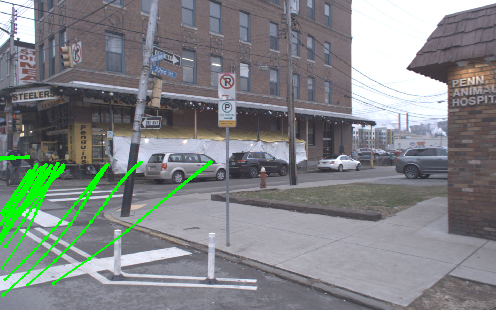}}&
         \raisebox{-.5\height}{\includegraphics[width=0.20\linewidth, height=0.13\linewidth]{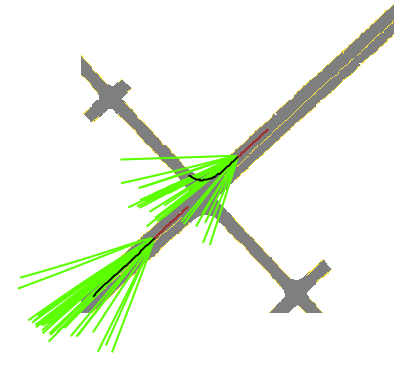}}
         \\
         S-GAN\cite{social_gan}&
         \raisebox{-.5\height}{\includegraphics[width=0.20\linewidth]{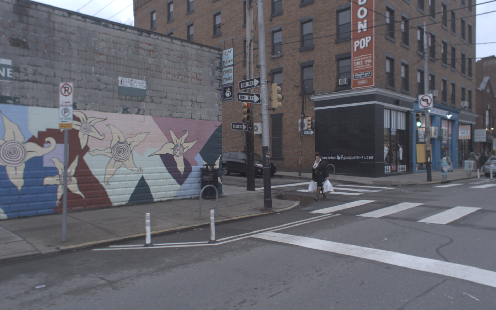}}&
         \raisebox{-.5\height}{\includegraphics[width=0.20\linewidth]{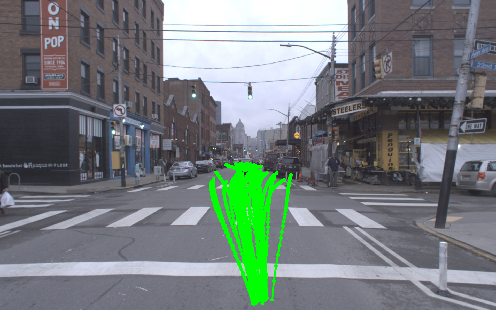}}&
         \raisebox{-.5\height}{\includegraphics[width=0.20\linewidth]{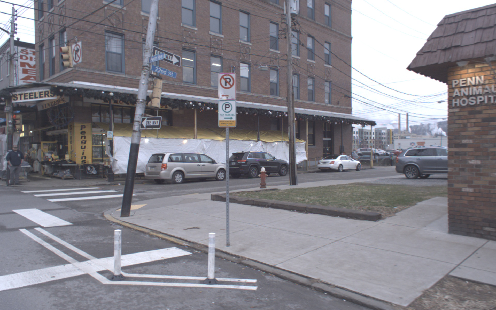}}&
         \raisebox{-.5\height}{\includegraphics[width=0.20\linewidth, height=0.13\linewidth]{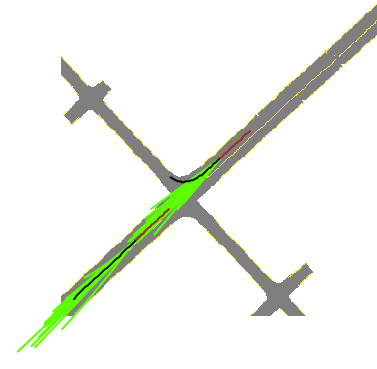}}
         \\
         SMART &
         \raisebox{-.5\height}{\includegraphics[width=0.20\linewidth]{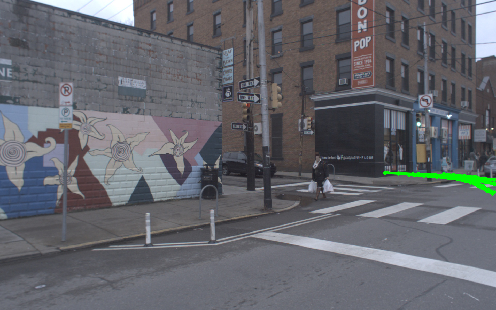}}&  
         \raisebox{-.5\height}{\includegraphics[width=0.20\linewidth]{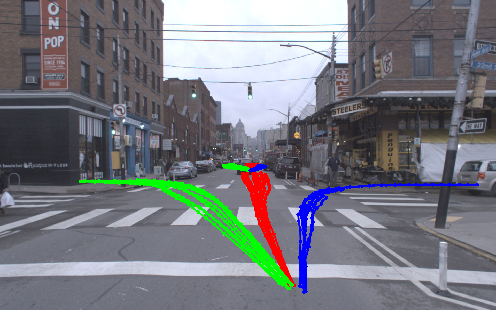}}&
         \raisebox{-.5\height}{\includegraphics[width=0.20\linewidth]{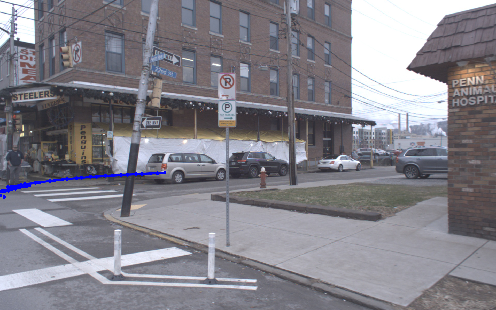}}&
         \raisebox{-.5\height}{\includegraphics[width=0.20\linewidth,height=0.13\linewidth]{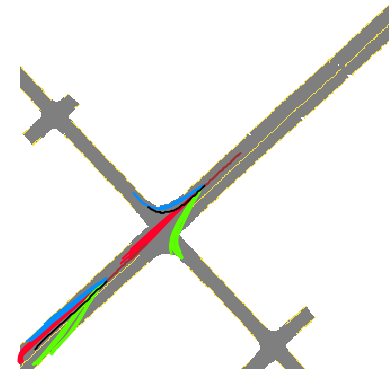}}\\
          
    \end{tabular}
    
    \caption{Qualitative comparison of SMART with other baseline methods. The figure depicts predicted outputs plotted on left, center, right camera images and top view images of the scene. Note that for SMART, red, blue and green trajectories show outputs with different trajectory labels that captures different modes in the output.}
    \label{fig:supp_results_2}
\end{figure}

Figures \ref{fig:supp_results} and \ref{fig:supp_results_2} show qualitative comparisons of the proposed SMART method with other baselines.

\end{document}